%% file: main.tex
\documentclass[10pt,twocolumn,letterpaper]{article}

\usepackage[pagenumbers]{iccv} 


\usepackage{array}
\newcommand{\PreserveBackslash}[1]{\let\temp=\\#1\let\\=\temp}
\newcolumntype{C}[1]{>{\PreserveBackslash\centering}p{#1}}
\newcolumntype{R}[1]{>{\PreserveBackslash\raggedleft}p{#1}}
\newcolumntype{L}[1]{>{\PreserveBackslash\raggedright}p{#1}}
\usepackage{multicol,lipsum}

\definecolor{iccvblue}{rgb}{0.21,0.49,0.74}
\usepackage[pagebackref,breaklinks,colorlinks,allcolors=iccvblue]{hyperref}
\usepackage{multirow}
\newcommand*\rot{\rotatebox{90}}
\usepackage[accsupp]{axessibility}  

\def\paperID{2} 
\def\confName{ICCV}
\def\confYear{2025}

\title{WildSAT: Learning Satellite Image Representations from Wildlife Observations}

\author{
Rangel Daroya$^1$ \quad
Elijah Cole$^2$ \quad 
Oisin Mac Aodha$^3$ \quad
Grant Van Horn$^1$ \quad
Subhransu Maji$^1$ \\
$^1$University of Massachusetts, Amherst \quad
$^2$GenBio AI \quad 
$^3$University of Edinburgh
}

\begin{document}

\twocolumn[{%
\renewcommand\twocolumn[1][]{#1}%
\maketitle\begin{center}
    \centering
    \captionsetup{type=figure}
    \includegraphics[scale=0.29]{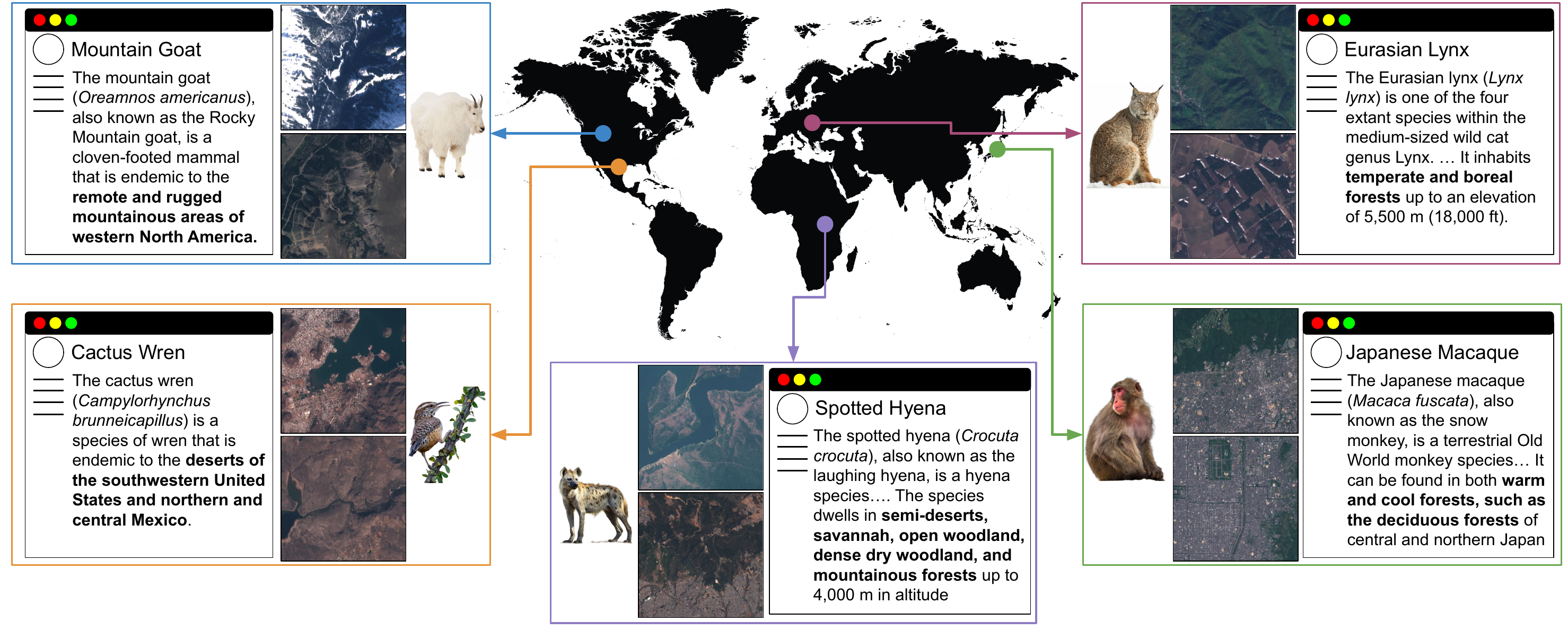}
    \vspace{-10pt}
    \captionof{figure}{\textbf{Wildlife observations can provide valuable supervision for learning satellite image representations.} Known wildlife locations derived from human observations, coupled with descriptive information on species range, habitat, and other ecological attributes on Wikipedia, serve as a rich source of contextual information for satellite imagery. Our \textbf{WildSAT} approach leverages these additional data sources to (i) learn robust satellite image representations for downstream tasks, and (ii) complement and further improve existing models using continual pre-training.
    }
    \label{fig:motivation}
     \vspace{4pt}
\end{center}%
}]

\begin{abstract}
Species distributions encode valuable ecological and environmental information, yet their potential for guiding representation learning in remote sensing remains underexplored.
We introduce WildSAT, which pairs satellite images with millions of geo-tagged wildlife observations readily-available on citizen science platforms.
WildSAT employs a contrastive learning approach that jointly leverages satellite images, species occurrence maps, and textual habitat descriptions to train or fine-tune models.
This approach significantly improves performance on diverse satellite image recognition tasks, outperforming both ImageNet-pretrained models and satellite-specific baselines.
Additionally, by aligning visual and textual information, WildSAT enables zero-shot retrieval, allowing users to search geographic locations based on textual descriptions.
WildSAT surpasses recent cross-modal learning methods, including approaches that align satellite images with ground imagery or wildlife photos, demonstrating the advantages of our approach. Finally, we analyze the impact of key design choices and highlight the broad applicability of WildSAT to remote sensing and biodiversity monitoring.
\end{abstract}

\vspace{-   15pt}
\section{Introduction}
\label{sec:intro}

The growth in the number of satellites with imaging capabilities deployed over the past 50 years has provided an unprecedented ability to monitor the surface of the earth~\cite{zhu2017deep, yuan2020deep, kussul2017deep}. 
The image data derived from these remote sensors has been shown to be highly effective for diverse tasks such as estimating global tree canopy height~\cite{lang2023high, torres2022canopy}, detecting illegal fishing activity~\cite{paolo2022xview3, kurekin2019operational, geronimo2018mapping}, crop monitoring~\cite{karmakar2024crop, wu2015global, fan2022gnn}, disaster management~\cite{sun2020applications, voigt2007satellite, perez2024discount}, among others.  
Central to building computer vision models for these tasks is the need for mechanisms for learning effective representations from image data. 
As a result of the distribution shift between remote sensing imagery and web-sourced images, a large body of work has emerged exploring the merits and trade-offs between different sources of supervision. 

Direct supervision in the form of paired images and labels (\eg image tiles with labels denoting land cover type) can be prohibitively expensive to obtain at a global scale~\cite{helber2019eurosat}. 
To address this, there is growing interest to develop methods that learn remote sensing representations from self-supervision~\cite{manas2021seasonal_seco, mall2023change_caco, jakubik2310foundation_prithvi}, multiple paired modalities~\cite{taxabind2025, dhakal2024geobind,mall2024remote_graft}, or other auxiliary sources~\cite{vivanco2024geoclip, dollinger2024sat}. 
A useful supervision source needs to be globally distributed, correlated with the local landscape as viewed from an image, and able to discriminate regions at a fine spatial scale.

A promising auxiliary supervision source is provided by locations where different species of plants and animals can be found around the world. 
For example, \textit{Mountain Goats (Oreamnos americanus)} are found in rugged mountainous areas, while habitat specialists like the \textit{Cactus Wren (Campylorhynchus brunneicapillus)} are typically found in deserts nesting in spiny cacti (\cref{fig:motivation}). 
Species location data offer a rich source of supervision, reflecting the local natural environment around each observation.
It is also readily available from citizen science platforms such as iNaturalist~\cite{inatWeb} and eBird~\cite{sullivan2009ebird} which host hundreds of millions of wildlife observations. 
While species location data has improved fine-grained species classification~\cite{berg2014birdsnap,chu2019geo,mac2019presence}, its potential for learning remote sensing representations remains unclear. Prior works have largely relied on anthropogenic labels (\eg human-made features like roads, buildings, industrial areas) to learn satellite image representations~\cite{uzkent2019learning, mall2024remote_graft, liu2024remoteclip}, whereas we explore using wildlife observations as a complementary and potentially valuable signal.

We introduce a new approach that uses signals derived from species location observations. 
We take inspiration from recent work that attempt to fuse multi-modal ecological data and remote sensing imagery into a shared common embedding space~\cite{sastry2024birdsat,huynh2024contrastive,taxabind2025}. 
WildSAT uses a contrastive learning objective to align satellite image, text, and location based on species observation data, bringing embeddings from the same area closer together and pushing those from different areas further apart.
Through this method, we utilize information about the preferred habitats of species to improve satellite image representations.

We make the following contributions:  
(i) We introduce \textbf{WildSAT}, a new approach to learning remote sensing representations using species observation locations as a supervisory signal.
(ii) We show WildSAT-derived representations are competitive with state-of-the-art satellite representations, while enabling zero-shot satellite image retrieval.
(iii) We present a thorough evaluation that highlights WildSAT-derived representations not only outperform but also complement existing methods focused on anthropogenic labels by incorporating wildlife information.
(iv) We perform ablation studies to show the impact of each component of our approach, and show WildSAT outperforms recent cross-modal methods like GRAFT~\cite{mall2024remote_graft} and TaxaBind~\cite{taxabind2025}. The code and dataset are available at \url{https://github.com/cvl-umass/wildsat}.

\section{Related Work}
\label{sec:relatedwork}
Previous works learn satellite image representations by training on large-scale remote sensing datasets from programs like Landsat~\cite{masek2020landsat}, Sentinel~\cite{immitzer2016first, esa2022sentinel}, or NAIP~\cite{maxwell2017land}.
These methods range from using self-supervised~\cite{manas2021seasonal_seco, cong2022satmae, jakubik2310foundation_prithvi}, supervised~\cite{bastani2023satlaspretrain, neumann2019domain, sumbul2019bigearthnet}, and cross-modal~\cite{nedungadi2024mmearth, taxabind2025, dhakal2024geobind, mall2024remote_graft, picek2024geoplant, Guo_2024_CVPR_skysense} learning to learn rich image representations for downstream satellite-based tasks.

\noindent\textbf{Self-supervised learning.}
These methods learn representations by taking advantage of spatio-temporal invariance or by predicting missing image patches from satellite images. 
SeCo~\cite{manas2021seasonal_seco} collects data from the same location across different seasons and uses a contrastive objective to force the image embeddings of samples to be closer if they are from the same location, and farther otherwise. 
Other works~\cite{mall2023change_caco, zheng2024changen2} extend this by selecting points in time based on the level of similarity between satellite images, or by synthetically generating images that are variants of the same location. 
Vision transformers trained using masked autoencoders~\cite{he2022masked, jakubik2310foundation_prithvi, cong2022satmae} have also been adopted owing to their success in learning natural image representations.
A notable example is Prithvi~\cite{jakubik2310foundation_prithvi}, a transformer network with 100 million parameters pretrained on 1TB of satellite imagery from around the world, achieving strong performance across a variety of Earth observation tasks.

\noindent\textbf{Supervised learning.}
These methods leverage labels from tasks like object detection~\cite{xia2018dota}, instance segmentation~\cite{waqas2019isaid}, and classification~\cite{helber2019eurosat, sumbul2019bigearthnet} in the satellite domain. Some works focus solely on image classification~\cite{xia2018dota, waqas2019isaid, xia2017aid, sumbul2019bigearthnet, neumann2019domain}, while others learn from multiple label types. For example, SatlasPretrain~\cite{bastani2023satlaspretrain} curated a large dataset with over 300 million labels across 137 categories, using domain experts, crowdsourced workers, and publicly available datasets (\eg OpenStreetMap~\cite{haklay2008openstreetmap}, WorldCover~\cite{van2021esa_worldcover}). It is a unified model jointly trained for multiple tasks (\eg segmentation, object detection) and showed improved performance on various downstream remote-sensing tasks, significantly outperforming ImageNet-pretrained models and other baselines.

\begin{figure*}[!t]
    \begin{center}
    \includegraphics[scale=0.65]{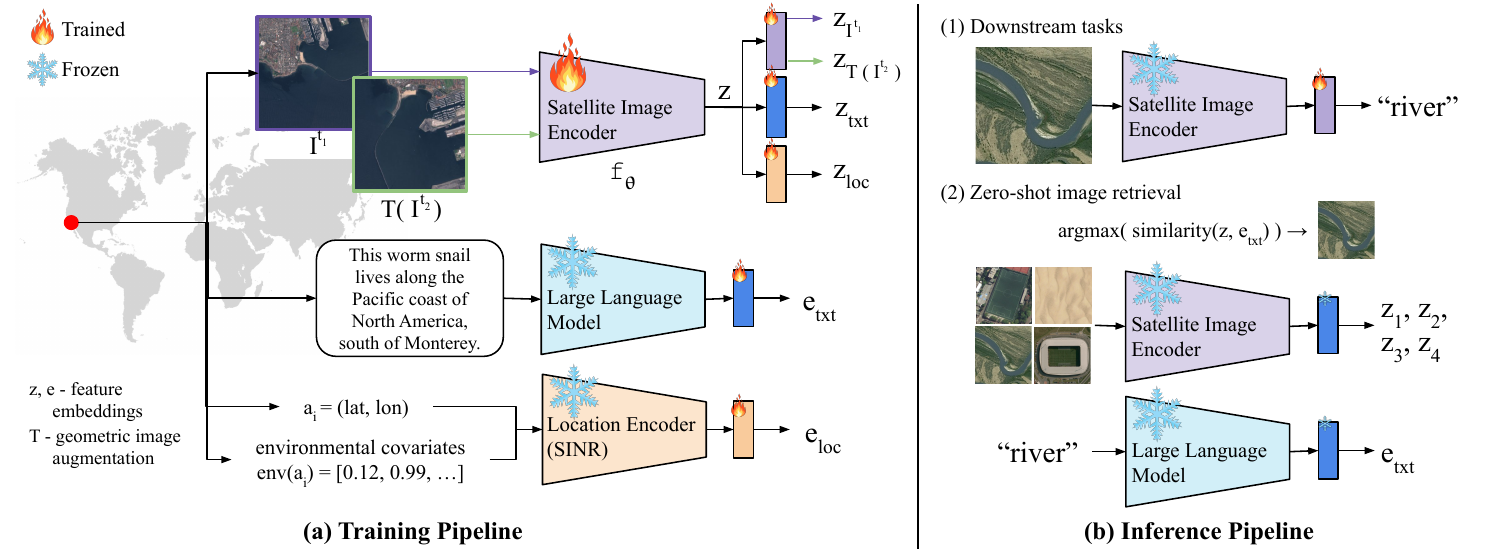}
    \end{center}
    \vspace{-18pt}
    \caption{\textbf{Architecture for training and evaluating the satellite image encoder}. (a) The training pipeline uses the location of a species, the satellite images at those locations, the environmental covariates, and the Wikipedia text associated with the species. In addition to the alignment of image, text, and location modalities, the encoder is encouraged to learn additional image features by using temporal and geometric image transformations on the input satellite image. (b) Downstream tasks use the frozen satellite image encoder with an additional trainable layer (or layers). Alternatively, the predicted image embeddings can be used for zero-shot retrieval via text queries. 
    }
    \label{fig:architecture}
    \vspace{-12pt}
\end{figure*}

\noindent\textbf{Cross-modal learning.}
Recently, other works have explored adding other modalities while training on satellite images~\cite{Guo_2024_CVPR_skysense,klemmer2023satclip, sastry2024birdsat, vivanco2024geoclip,mall2024remote_graft, huynh2024contrastive,taxabind2025, dhakal2024geobind, khanal2023learning}.
A common approach uses geo-tagged images and pre-trained image-text encoders like CLIP~\cite{radford2021learning_clip}, aligning new modalities to their embedding space using contrastive learning~\cite{klemmer2023satclip, mall2024remote_graft, dhakal2024geobind, vivanco2024geoclip, jain2024aligning, liu2024remoteclip}. 
This strategy has been used for various tasks: satellite image localization in GeoCLIP~\cite{vivanco2024geoclip}, bird species classification and mapping in BirdSAT~\cite{sastry2024birdsat}, and improving plant species image representations in CRISP~\cite{huynh2024contrastive}.
Models like GRAFT~\cite{mall2024remote_graft}, TaxaBind~\cite{taxabind2025}, RemoteCLIP~\cite{liu2024remoteclip}, and GeoBind~\cite{dhakal2024geobind} align multiple modalities at the same time for cross-modal retrieval and zero-shot tasks. 
\citet{zermatten2025ecowikirs} have also demonstrated the benefits of aligning satellite imagery with species observation data for zero-shot classification.
TaxaBind was the first to use species geographic locations and satellite imagery, but it focuses on ecological tasks rather than satellite image tasks. We also expand on their approach by leveraging open-source Wikipedia text instead of taxonomic hierarchy data, offering a more diverse supervision for satellite image representations. 
Beyond contrastive learning, other methods use supervised learning to fuse embeddings of different modalities for predicting species range maps and encounter rates~\cite{dollinger2024sat, hamilton2024combining, teng2024satbird}.
Most similar to our work is WikiSatNet~\cite{uzkent2019learning} which uses location-aligned Wikipedia articles and satellite images to improve satellite image representations.
However, while WikiSatNet and previous works~\cite{mall2024remote_graft, uzkent2019learning, liu2024remoteclip, klemmer2023satclip} primarily focus on anthropogenic data, we explore the impact of wildlife observations.
Specifically, we investigate how species distributions---capturing habitat preferences, climate, and environmental factors---can serve as powerful signals. 

While previous work has focused on improving species distribution modeling~\cite{dollinger2024sat, teng2024satbird, sastry2024birdsat, lorieul2022overview} or fine-grained image classification~\cite{taxabind2025, dhakal2024geobind, mai2023csp} using satellite images, our work improves satellite image representations using wildlife observations. Our experiments show that both randomly initialized models and strong baselines, such as Prithvi~\cite{jakubik2310foundation_prithvi}, SatlasNet~\cite{bastani2023satlaspretrain}, and SeCo~\cite{manas2021seasonal_seco}, benefit from this supervision on a wide range of satellite image tasks (\cref{fig:linear-probing}, \cref{table:segmentation-results-iou}).

\section{Method}
\label{sec:method}

We define the problem as follows: given an image encoder $f_\theta : \mathbf{I} \rightarrow \mathbf{z}$ with parameters $\theta$, we want to find an optimal set of parameters $\theta^*$ that improves the performance of $f$ on various remote sensing tasks through a robust satellite image feature representation $\mathbf{z}$.
It takes an image $\mathbf{I} \in \mathbb{R}^{W\times H\times3}$ as input and outputs an embedding $\mathbf{z} \in \mathbb{R}^{d}$.
We propose to optimize $\theta$ using our WildSAT framework using data consisting of satellite images, locations, environmental covariates, and text. We hypothesize that leveraging known environmental context around each species observation (\cref{fig:motivation}) allows for more effective optimization of model parameters.

\subsection{WildSAT}
To supplement satellite images, we take advantage of additional modalities that naturally align based on the distribution of species throughout the globe.
Information on species habitat can provide a rich source of supervision for improving satellite image representations, and we describe how to leverage this through our proposed WildSAT framework.
\cref{fig:architecture} shows the overall architecture used to train a satellite image encoder $f_\theta$. 
The encoder $f$ can be any architecture (\eg a ResNet50~\cite{he2016deep}, ViT-B/16~\cite{dosovitskiy2020image}, \etc). 
The initial parameters $\theta$ can be randomly initialized, pre-trained on a different domain (\eg ImageNet~\cite{deng2009imagenet}), or pre-trained on a related dataset (\eg SatlasPretrain~\cite{bastani2023satlaspretrain}).
The output embedding $\mathbf{z}$ can be used for downstream remote sensing tasks such as classification and zero-shot image retrieval.

WildSAT aims to improve the model by training on additional modalities related to species observation data.
To incorporate other modalities, we use pre-trained models, \eg  SINR~\cite{cole2023spatial_sinr} for location and GritLM~\cite{muennighoff2024generative} for text.

\noindent \textbf{Location encoder.} SINR~\cite{cole2023spatial_sinr} is mainly used to predict the presence and absence of species at a location by training on large collections of species observation data. 
It takes location $\mathbf{a}_i=(lat,lon) \in \mathbb{R}^2$ and, optionally, the corresponding environmental covariates (\eg weather and climate data) to produce a location embedding in $\mathbb{R}^{256}$. 

\noindent \textbf{Large language model}. GritLM~\cite{muennighoff2024generative} is a large language model (LLM) that outputs a fixed-length embedding in $\mathbb{R}^{4096}$ given a text input. It is trained to handle text of arbitrary length, making it suitable for longer and varying lengths of text obtained from sources such as Wikipedia~\cite{wikipedia}. 

\noindent \textbf{Satellite image encoder}. Given an initial image encoder $f$, we add three sets of linear layers to predict embeddings for  images~($\mathbf{z}_{I^t}$), text~($\mathbf{z}_{\text{txt}}$), and locations~($\mathbf{z}_{\text{loc}}$). 
Similarly, both the pre-trained LLM (GritLM) and location encoder (SINR) have an added trainable linear layer to project their respective feature embeddings to $\mathbb{R}^d$ as $\mathbf{e}_{\text{txt}}$ and $\mathbf{e}_{\text{loc}}$, respectively. In addition to text and location, we also fine-tune the model on image-specific features by forcing embeddings of similar satellite images to be close to each another. 
For two satellite images $\mathbf{I}^{t_1}$ and $\mathbf{I}^{t_2}$ taken at the same location but at different times, their corresponding feature representations should be similar.
We also apply geometric augmentations $T$ such as flipping and random cropping on the latter image such that $f(\mathbf{I}^{t_1}) \approx f(T(\mathbf{I}^{t_2}))$.
Doing this encourages the model to learn meaningful image features by distinguishing between similar and dissimilar images while refining the same features through text, location, and environment.

\subsection{Training}
Our framework uses a contrastive learning objective to improve satellite image encoder embeddings. We jointly optimize the parameters of the model $f_\theta$ and the additional linear layers through the training objective in Eqn.~\ref{eq:optimization}. These loss terms correspond to a contrastive objective over image embeddings~($\mathcal{L}_{\text{img}}$), text embeddings~($\mathcal{L}_{\text{txt}}$), and location embeddings~($\mathcal{L}_{\text{loc}}$) of $f$. The embeddings $\mathbf{z}_{\mathbf{I}^{t}},\mathbf{z}_{\text{txt}},\mathbf{z}_{\text{loc}}$ are linear projections of the image embedding for each of the modalities (see \cref{fig:architecture}). Eqn~\ref{eq:optimization} shows the objective.
\vspace{-1mm}
\begin{equation}
    \min_\theta \left[ \underbrace{\mathcal{L}(\mathbf{Z}_{\mathbf{I}^{t_1}},\mathbf{Z}_{T\left(\mathbf{I} ^ {t_2} \right)})}_{\mathcal{L}_{\text{img}}} + \underbrace{\mathcal{L}(\mathbf{Z}_{\text{txt}},\mathbf{E}_{\text{txt}})}_{\mathcal{L}_{\text{txt}}} + \underbrace{\mathcal{L}(\mathbf{Z}_{\text{loc}},\mathbf{E}_{\text{loc}})}_{\mathcal{L}_{\text{loc}}} \right]\label{eq:optimization}
\end{equation}

We compute the distance between two sets of embeddings $\mathbf{Z}$ and $\mathbf{E}$ using a minibatch of $n$ samples with the $i$-th embedding in $\mathbf{Z}$ aligned with the $i$-th embedding of $\mathbf{E}$.
That is, given two sets of embeddings $\mathbf{Z}=\{\mathbf{z}_1,\ldots,\mathbf{z}_n\}$ and $\mathbf{E}=\{\mathbf{e}_1,\ldots,\mathbf{e}_n\}$, the embedding $\mathbf{z}_i$ is matched to $\mathbf{e}_i$ against other embeddings $\mathbf{e}_{1, \ldots, n}$ using the loss in Eqn.~\ref{eq:loss-bet-vectors}.

{
\begin{equation}
    \mathcal{L}(\mathbf{Z},\mathbf{E}) = \frac{1}{2n} \sum_{i=1}^n \big( \mathcal{L}_{con} (\mathbf{z}_i, \mathbf{e}_{1,\ldots,n}) + \mathcal{L}_{con} (\mathbf{e}_{i}, \mathbf{z}_{1,\ldots,n}) \big)
    \label{eq:loss-bet-vectors}
\end{equation}
}

Based on InfoNCE loss~\cite{oord2018representation_contrastive, radford2021learning_clip}, Eqn.~\ref{eq:contrastive-loss} matches the embedding $\mathbf{z}_i$ with the corresponding embedding $\mathbf{e}_i$ by minimizing the distance between them with respect to the other embeddings in $\mathbf{E}$, with temperature hyperparameter $\tau$.
\begin{equation}
    \mathcal{L}_{con}(\mathbf{z}_i, \mathbf{e}_{1,\ldots,n}) = -\log \frac{\exp \left( \mathbf{z}_i \cdot \mathbf{e}_i / \tau\right)}{
                    \sum_{j=1}^n \exp \left( \mathbf{z}_i \cdot \mathbf{e}_j / \tau \right)
                }
    \label{eq:contrastive-loss}
\end{equation}

\subsection{Implementation Details}
During training, we fine-tune all satellite image encoders and added linear layers on the species observation dataset using Eqn.~\ref{eq:optimization}. For models pre-trained on out-of-domain datasets (\eg ImageNet1K~\cite{deng2009imagenet}), we apply parameter-efficient fine-tuning (PEFT) tailored to each architecture:
ResNet50 uses scale and shift fine-tuning~\cite{frankle2021training,lian2022scaling}, tuning only BatchNorm parameters, while ViT and Swin~\cite{liu2021swin} use DoRa~\cite{mao2024dora} on the attention layers.
These techniques enable gradual parameter updates, allowing models to learn new satellite image features without forgetting those from their original domain.
For a randomly initialized model or a model pre-trained in the same domain (\ie satellite images), we fine-tune all parameters.

\section{Dataset}
To train the model, we combine images, text, location, and environmental covariates from publicly available datasets~\cite{cole2023spatial_sinr, hamilton2024combining, fick2017worldclim, wikipedia, esa2022sentinel}.
For a given species, we obtain its corresponding observation data through iNaturalist~\cite{van2018inaturalist}, and a text description of its preferred habitat from its corresponding Wikipedia~\cite{wikipedia} page (\cref{fig:motivation}).
Satellite images are then retrieved based on the species observation locations. 
We describe each component of the dataset below.

\noindent\textbf{Location observation data.} 
iNaturalist~\cite{van2018inaturalist} observations consist of latitude and longitude values denoting the locations where a species has been observed.
We use the dataset from SINR~\cite{cole2023spatial_sinr} that contains 35.5 million observations of 47,375 different species.
It is composed of $\{(\mathbf{a}_i, b_i)\}_{i=1}^N$ pairs where $\mathbf{a}_i=(lat,lon)$ is the location and $b_i\in\{s_1, s_2, \dots, s_M\}$ is an integer encoding the species ID of the observed species.
The locations of each species are used as the basis for matching other sources of data such as environmental covariates, satellite images, and text.

\noindent\textbf{Environmental covariates.}
Environmental covariates are obtained from WorldClim2~\cite{fick2017worldclim} for a location $\mathbf{a}_i$.
The data aggregate temperature and precipitation values and are often used for ecological applications.
It returns a vector for each location $\text{env}(\mathbf{a}_i) \in \mathbb{R}^{20}$.
We use five-minute resolution data and bilinearly interpolate between data points to get values for specific locations.

\noindent\textbf{Satellite images.} Sentinel-2 satellite images are used with a resolution of ten meters per pixel and a size of 512$\times$512. The satellite provides 13 spectral bands at different resolutions and has a revisit frequency of five days near the equator. 
We remove satellite images with significant cloud cover. Satellite images taken at the same locations but at a different time are collected as a form of augmentation for training. A total of 305,689 satellite images are collected~\cite{esa2022sentinel, bastani2023satlaspretrain}.

\noindent\textbf{Text.} For each species, we use readily available text data. 
The text is taken from Wikipedia~\cite{wikipedia} as in LE-SINR~\cite{hamilton2024combining}. The corresponding page typically has several sections describing different aspects of the species such as its habitat, range, behavior, taxonomy, \etc 
Each section is processed separately so that a text embedding corresponds to a single section of the Wikipedia text.
Similar to~\cite{hamilton2024combining}, we remove sections that do not provide information about the species (\eg references, bibliography, links).
The final text dataset contains 127,484 sections from 37,889 species.

For a single satellite image-species location match, there could be multiple text embeddings that correspond to the multiple sections of the species Wikipedia~\cite{wikipedia} page. 
Taking these into account, there are a total of 980,376 training samples with location, satellite image, and text. 
During training, we randomly sample one section of text for every satellite image-species location match, resulting in effectively 134,890 iterations per epoch. 
We show the spatial distribution of the data in \cref{fig:data-distribution} (Appendix), and provide more training details in the Appendix. 

\section{Experiments}
\label{sec:experiments}
We evaluate the representations learned by \textbf{WildSAT} via linear probing experiments. 
Starting with different models and different parameter initializations (either random or pre-trained), we evaluate the performance before and after fine-tuning.
When probing for each downstream dataset, the trained satellite image encoder is frozen and a randomly initialized decoder is added (\cref{fig:architecture}b.1). 
For all tasks except segmentation, a linear layer is used for the decoder.
Segmentation tasks use a convolutional-based decoder with a U-Net architecture~\cite{ronneberger2015u}.
Only the decoder is trained for each downstream task to assess the impact of the image embedding $\mathbf{z}$ without diluting its representation.

\subsection{Satellite Image Classification and Segmentation}
Nine remote sensing classification datasets were used as downstream tasks to evaluate the performance of the image embeddings on classification and segmentation.
For classification, we evaluate on UCM~\cite{yang2010bag_ucmdataset}, AID~\cite{xia2017aid}, RESISC45~\cite{cheng2017remote_resisc45dataset}, FMoW~\cite{christie2018functional_fmow}, EuroSAT~\cite{helber2019eurosat}, So2Sat20k~\cite{zhu2020so2sat, lacoste2024geobench}, and BigEarthNet20k (BEN20k)~\cite{sumbul2019bigearthnet, lacoste2024geobench}. For segmentation, we evaluate on Cashew1k~\cite{yin2023mapping} and SAcrop3k~\cite{sacrop}.
Classes vary from man-made structures (\eg airplanes, buildings) to land type (\eg forest, vegetation). 
Of the seven datasets, one (BEN20k) is multi-label, and the rest are single-label. 
We report accuracy for single-label classification datasets, micro-F1 for BEN20k, and IoU for both multi-class segmentation datasets.
Datasets are described in the Appendix.

\begin{figure*}[t]
    \centering
    \begin{minipage}{0.83\textwidth}
        \centering
        \includegraphics[scale=0.52]{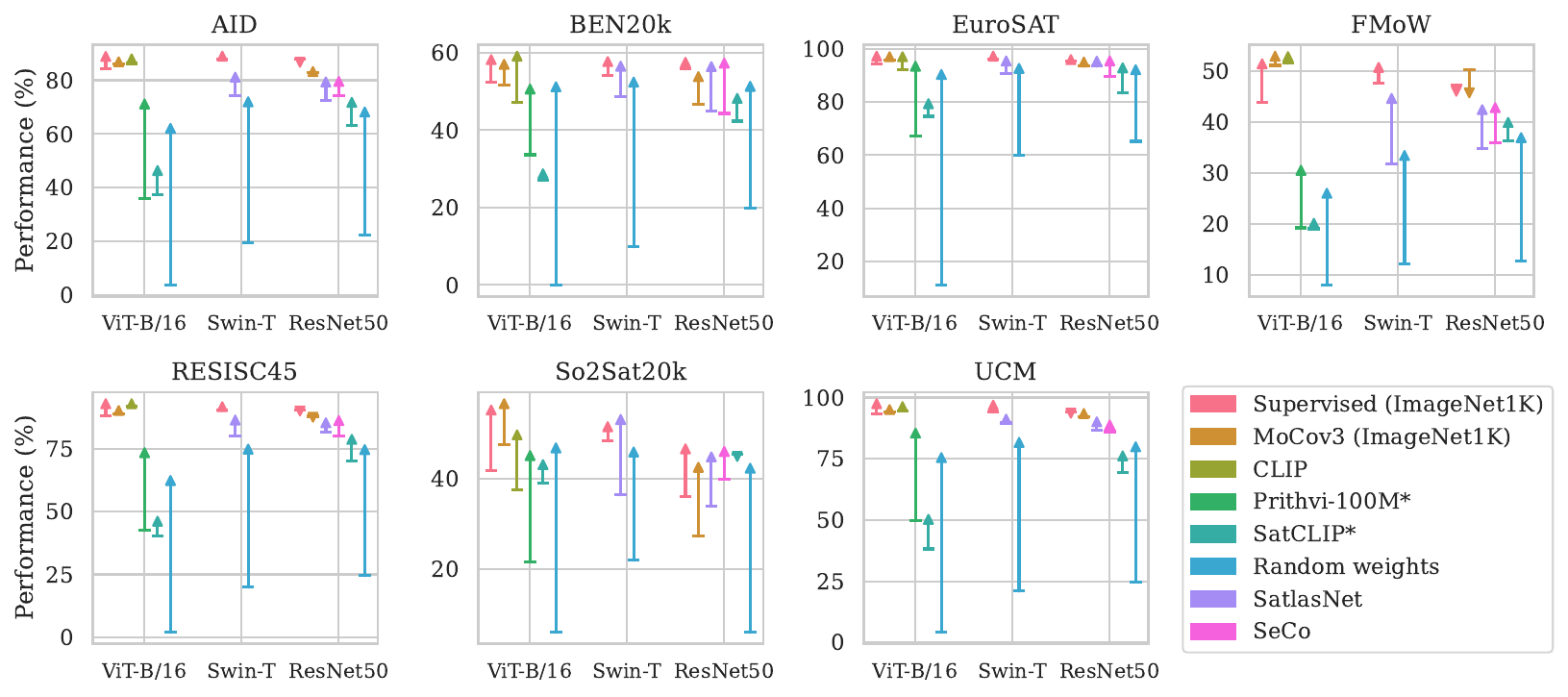}
        \label{fig:linear-probing}
    \end{minipage}%
    \hfill
    \begin{minipage}{0.17\textwidth}
        \centering
        \footnotesize
        \setlength{\tabcolsep}{2pt}
        \resizebox{0.99\linewidth}{!}{
        \begin{tabular}{l |r r }
            \toprule	
            & \multicolumn{2}{c}{Average} \\
            & \multicolumn{2}{c}{(w/ random)} \\
                \midrule
            Dataset & Base & +WS \\
                \midrule
            AID~\cite{xia2017aid} & 61.2  & \textbf{77.0}  \\
            BEN20k~\cite{sumbul2019bigearthnet, lacoste2024geobench} & 38.5  & \textbf{53.1}  \\
            EuroSAT~\cite{helber2019eurosat} & 80.2  & \textbf{93.8}  \\
            FMoW~\cite{christie2018functional_fmow} & 33.4  & \textbf{41.1}  \\
            RESISC45~\cite{cheng2017remote_resisc45dataset} & 65.3  & \textbf{81.0}  \\
            So2Sat20k~\cite{zhu2020so2sat, lacoste2024geobench} & 32.6  & \textbf{47.6}  \\
            UCM~\cite{yang2010bag_ucmdataset} & 68.8  & \textbf{86.1}  \\
            \bottomrule
            \toprule
            & \multicolumn{2}{c}{Average} \\
            & \multicolumn{2}{c}{(no random)} \\
                \midrule
            Dataset & Base & +WS \\
                \midrule
            AID~\cite{xia2017aid} & 72.7  & \textbf{79.4}  \\
            BEN20k~\cite{sumbul2019bigearthnet, lacoste2024geobench} & 45.7  & \textbf{53.4}  \\
            EuroSAT~\cite{helber2019eurosat} & 88.9  & \textbf{94.3}  \\
            FMoW~\cite{christie2018functional_fmow} & 39.0  & \textbf{43.3}  \\
            RESISC45~\cite{cheng2017remote_resisc45dataset} & 77.8  & \textbf{83.5}  \\
            So2Sat20k~\cite{zhu2020so2sat, lacoste2024geobench} & 37.9  & \textbf{48.2}  \\
            UCM~\cite{yang2010bag_ucmdataset} & 81.8  & \textbf{87.9}  \\

            \bottomrule
            
        \end{tabular}        
        }
        \label{table:overall_results}
    \end{minipage}
    \vspace{-3mm}
    \caption{\textbf{Linear probing performance improvement on seven downstream datasets without (Base) and with WildSAT (+WS) fine-tuning}. Accuracy is visualized for all dataset plots except BEN20k that visualizes micro F1 score. For each architecture and pre-training combination, the horizontal line marker represents the performance of the original model, while the triangle marker indicates performance after additional training on species observation data (WildSAT). The tables on the right summarize average performance across all seven datasets: the top table includes models with random weights, and the bottom table excludes them. Across the board, fine-tuning with species observation data leads to notable performance gains over most base models. We include the raw numbers in \cref{table:overall-results} (Appendix). *Both Prithvi-100M and SatCLIP are pre-trained with multispectral images, but for consistency across downstream datasets and models, only RGB bands are used here. We show that WildSAT also improves on multispectral images in \cref{table:multispectral-results} (Appendix). }
    \label{fig:linear-probing}
    \vspace{-5mm}
\end{figure*}

\subsection{Bird Species Encounter Rate Prediction}
We further evaluate our satellite image representations by predicting species encounter rates from satellite images. 
SatBird~\cite{teng2024satbird} introduces a benchmark for predicting bird species encounter rates for an area given the satellite image. 
Encounter rates for a location are given as a vector $\mathbf{r} \in [0,1]^S$ for $S$ species, where the $i$-th element is the probability of observing species $i$ at that location. 
The benchmark includes three subsets: (1) USA summer, (2) USA winter, and (3) Kenya.
Models are evaluated using top-$k$ accuracy, which compares the species with the top $k$ predicted encounter rates with the actual species observed in an area. $k$ is the actual number of species present.

\subsection{Base Models}
Base models refer to the different pre-trained encoders before we fine-tune with WildSAT.
We experiment on 12 pre-training methods spanning random initialization, in-domain pre-training, and out-of-domain pre-training. These cover different architectures ResNet50, Swin-T, Swin-B, ViT-B/16, and ViT-L/16 for a total of 20 base models.

\noindent\textbf{ImageNet}~\cite{deng2009imagenet} models are pre-trained with supervision on the updated ImageNet1K V2 dataset~\cite{recht2019imagenet}. Previous works use V1 of the dataset~\cite{deng2009imagenet}, but we opted for the updated version which improves performance on the ImageNet benchmark by 3-4\%. We include results on the ImageNet1K V1 pre-trained models in \cref{table:addtl-classification-results} (Appendix).

\noindent\textbf{MoCov3}~\cite{chen2021empirical_mocov3} models are pre-trained using self-supervision with InfoNCE~\cite{oord2018representation_contrastive} on ImageNet1K.

\noindent\textbf{CLIP}~\cite{radford2021learning_clip} uses a contrastive objective to train an image and text encoder on image-text pairs from the Internet.

\noindent\textbf{Prithvi-100M}~\cite{jakubik2310foundation_prithvi} models are pre-trained using self-supervision (MAE~\cite{he2022masked}) on the Harmonized Landsat Sentinel-2 (HLS)~\cite{claverie2018harmonized} data.

\noindent\textbf{SatCLIP}~\cite{klemmer2023satclip} is a self-supervised approach that uses paired location and satellite images from Sentinel-2. While the model has both a location and image encoder, we only use their image encoder for further fine-tuning and evaluation.

\noindent\textbf{SatlasNet}~\cite{bastani2023satlaspretrain} models are pre-trained using supervised learning on the SatlasPretrain dataset. The supervision spans a variety of label types ranging from segmentation to object detection and image classification.

\noindent\textbf{SeCo}~\cite{manas2021seasonal_seco} models are self-supervised on Setinel-2 images augmented in time.  

\noindent\textbf{TaxaBind}~\cite{taxabind2025} uses contrastive learning to learn a common embedding space for multiple modalities including satellite image, ground image, audio, and taxonomic text.

\noindent\textbf{GRAFT}~\cite{mall2024remote_graft} uses contrastive learning with the CLIP encoders to align ground images to satellite images and text.

\noindent\textbf{RemoteCLIP}~\cite{liu2024remoteclip} uses contrastive learning with CLIP to align various annotations of aerial and satellite images.

\noindent\textbf{SatMAE}~\cite{cong2022satmae} is a self-supervised model that uses MAE~\cite{he2022masked} on temporal multi-spectral satellite imagery.

\noindent\textbf{Random} denotes randomly initialized models.
\begin{table}[!t]
    \small
    \begin{center}
    \resizebox{0.85\linewidth}{!}{
    \begin{tabular}{l |r r | r r }
    \toprule	

	& \multicolumn{2}{c|}{Cashew1k~\cite{yin2023mapping}} & \multicolumn{2}{c}{SAcrop3k~\cite{sacrop}} \\
	& Base & +WS & Base & +WS \\
        \midrule
	ImageNet~\cite{deng2009imagenet} &  70.3\% & 70.6\% & 24.3\% & 25.0\% \\
	MoCov3~\cite{chen2021empirical_mocov3} & 71.4\% & 73.3\% & 22.9\% & 24.9\% \\
	SeCo~\cite{manas2021seasonal_seco} &  62.6\% & 73.3\% & 22.3\% & 22.8\% \\
	SatlasNet~\cite{bastani2023satlaspretrain} &  55.2\% & 71.0\% & 19.4\% & 20.5\% \\
	Random &  40.1\% & 72.6\% & 18.0\% & 20.3\% \\
	\midrule
        \midrule
	Average & 59.9\% & \textbf{72.2\%} & 21.4\% & \textbf{22.7\%} \\
     \bottomrule
     \end{tabular}
     }
     \vspace{-3mm}
     \caption{\textbf{Downstream satellite image segmentation results, reported using IoU, show WildSAT (+WS) improving on existing models}. Qualitative results are available in \cref{fig:segmentation} (Appendix).}
     \label{table:segmentation-results-iou}
     \end{center}
     \vspace{-6mm}
\end{table}

\section{Results and Discussion}

\label{sec:results}

\subsection{Classification and Segmentation Performance}
\cref{fig:linear-probing} and \cref{table:segmentation-results-iou} display the results on the 7 downstream classification datasets and 2 segmentation datasets, respectively, across 15 different architectures and pre-training methods. 
The addition of WildSAT improves 108 of the 115 settings with an overall average improvement ranging from 7.7\% to 17.4\% in the different datasets (4.3\% to 10.4\% without the randomly initialized models).

\begin{table}[!t]
    \setlength{\tabcolsep}{3pt}
    \small
    \begin{center}
    \resizebox{0.98\linewidth}{!}{
    \begin{tabular}{l | c c c c c}
    \toprule
    & TaxaBind & {GRAFT} & RemoteCLIP & {CLIP} & {WildSAT} \\
    & {\cite{taxabind2025}} & {\cite{mall2024remote_graft}} & \cite{liu2024remoteclip} & {\cite{radford2021learning_clip}} & {(Ours)} \\
    \midrule
    
    Average Performance & 59.8\% & 65.0\% & 71.0\% & 71.6\% & \textbf{76.6\%} \\

    \bottomrule
    \end{tabular}
    }
    \vspace{-3mm}
    \caption{\textbf{Average linear probing performance across all seven satellite image classification datasets of models based on CLIP.} TaxaBind, GRAFT, and RemoteCLIP fine-tune CLIP and use additional modalities (\eg text, satellite images, ground images) for cross-modal tasks. We show that fine-tuning CLIP with WildSAT improves performance on CLIP compared to other CLIP-based models. We include accuracy per dataset in \cref{table:clip-model-results} (Appendix).}
    \label{table:clip-model-results-summary}
    \end{center}
    \vspace{-8mm}
\end{table}

\noindent\textbf{WildSAT improves satellite image representations.} 
The results in \cref{fig:linear-probing} and \cref{table:segmentation-results-iou} highlight the performance improvements WildSAT contributes.
These improvements may be attributed to our use of diverse supervision—integrating images, text embeddings, and species data at scale.
This strategy ultimately helps in downstream tasks, particularly for both increasing true positive rates on classes related to habitats (\eg forests, deserts), while reducing false positives on the same types of classes.
We show how classes related to wildlife habitats improve performance in the Appendix.
Meanwhile, \cref{table:clip-model-results-summary} provides a comparison of WildSAT (CLIP ViT-B/16 as base) with TaxaBind~\cite{taxabind2025}, GRAFT~\cite{mall2024remote_graft}, and RemoteCLIP~\cite{liu2024remoteclip} all of which also fine-tune CLIP using cross-modal supervision. While the latter three methods show improvements on cross-modal tasks, their performance when linear probed on satellite imagery tasks suffers compared to CLIP, indicating a degree of ``forgetting" of the representations. In contrast, our method not only outperforms standard CLIP but also achieves the best overall results.

\begin{table}[!t]
    \setlength{\tabcolsep}{4pt}
    \small
    \begin{center}
    
    \resizebox{0.98\linewidth}{!}{
    \begin{tabular}{l | r r | r r | r r }
    \toprule
    & \multicolumn{2}{c|}{Kenya} & \multicolumn{2}{c|}{USA Summer} & \multicolumn{2}{c}{USA Winter} \\
    & Base & +WS & Base & +WS & Base & +WS \\
    \midrule
    SatlasNet~\cite{bastani2023satlaspretrain} & 23.90 & 24.40 & 48.59 & 50.03 & 54.02 & 55.01 \\
    SatMAE~\cite{cong2022satmae} & 23.66 & 23.75 & 46.09 & 48.69 & 52.40 & 53.86 \\
    TaxaBind~\cite{taxabind2025} & 23.83 & 24.20 & 48.31 & 49.87 & 53.76 & 55.00 \\
    \midrule
    \midrule
    Average & 23.80 & \textbf{24.12} & 47.66 & \textbf{49.53} & 53.39 & \textbf{54.62} \\
    \bottomrule
    \end{tabular}
    }
    \vspace{-6pt}
    \caption{\textbf{Top-$k$ accuracy for bird species encounter rate prediction on the SatBird~\cite{teng2024satbird} dataset.} Top-$k$ is defined as the accuracy of predicting the $k$ species present in an area. }
    \label{table:satbird-results}
    \end{center}
    \vspace{-9mm}
\end{table}
\noindent\textbf{Satellite pre-trained models see a larger boost in performance.} \cref{fig:linear-probing} demonstrates that training with WildSAT can improve performance by as much as 10\% on satellite pre-trained models such as SeCo and SatlasNet. 
While models pre-trained on ImageNet and MoCov3 also see performance improvements, we see less improvements on the AID, RESISC45, and FMoW datasets. 
This could be attributed to the three datasets having less categories related to species habitats (\eg storage tank, airport, church).
Additionally, since ImageNet and MoCov3 pre-trained models are already performing effectively, there is little room for improvement.
WildSAT is trained on a dataset that is specifically geared towards habitats and land characteristics. Thus, we see more improvements in So2Sat20k and BEN20k, which cover climate zones and land cover types.

\noindent\textbf{WildSAT reduces errors on habitat-related classes.} We show through a confusion matrix of So2Sat20k classes in \cref{fig:confusion-matrix-so2sat} (Appendix) that WildSAT achieves higher true positive counts across all types of classes (including anthropogenic classes) by reducing false positives on habitat-related categories (\eg `Scattered trees', `Dense trees'). This reduction in misclassifications contributes to overall performance gains, explaining the improvements observed across satellite image tasks.

\noindent\textbf{Larger performance gains on ViTs than ResNets.} While the ResNet50 ImageNet and MoCov3 models show smaller performance gains, their ViT counterparts show significantly larger improvements (\cref{fig:linear-probing}). Consistent with observations from previous work~\cite{radford2021learning_clip}, we see better performance with the addition of other modalities when using transformers. This could be attributed to the more flexible representations of ViTs, unlike CNNs that inherently incorporate a strong inductive bias from the use of convolution. Using attention layers in ViTs likely makes their embeddings more adaptable to other modalities such as text and location.

\begin{figure}[!t]
    \begin{center}
    \includegraphics[scale=0.25]{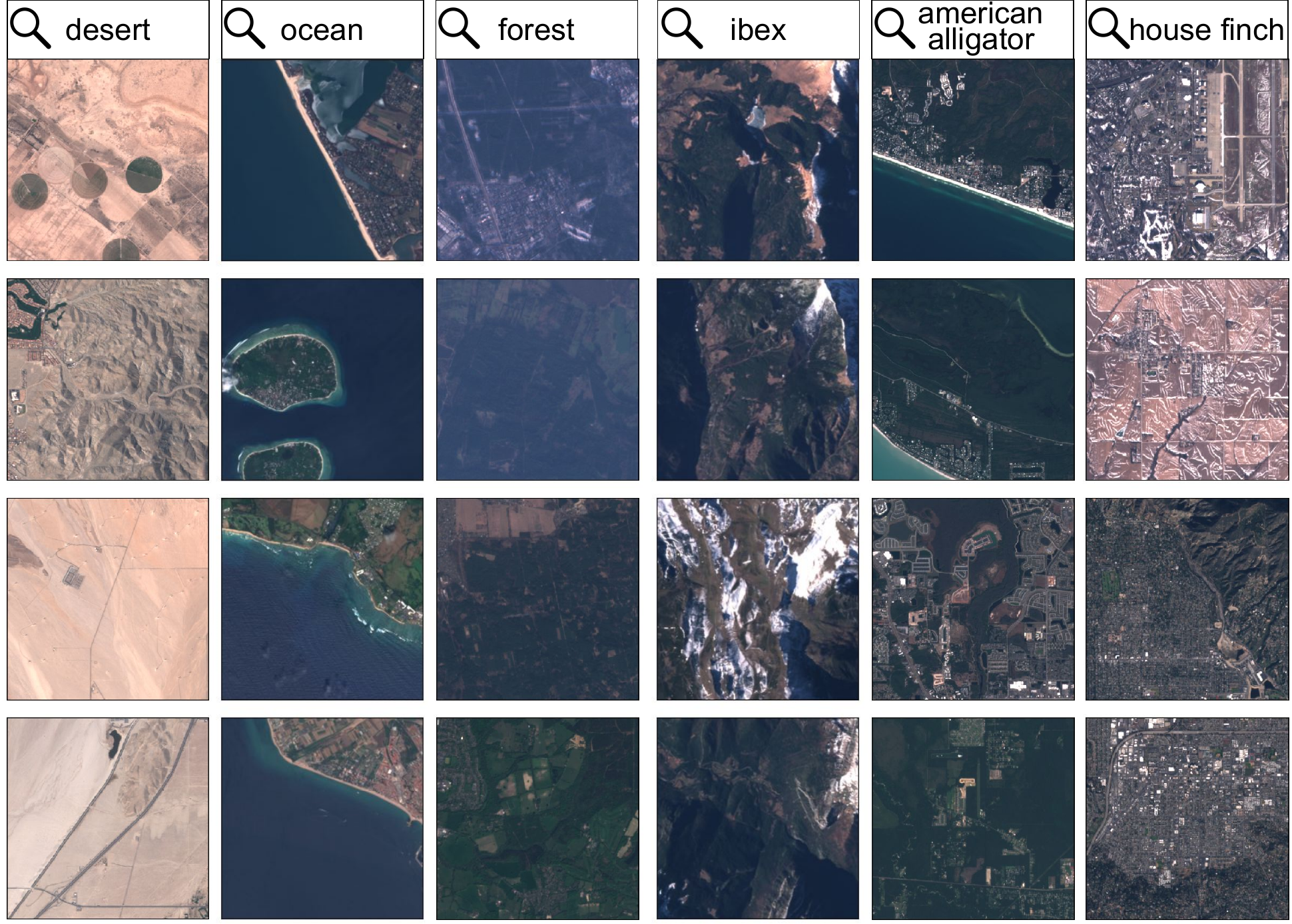}
    \end{center}
    \vspace{-16pt}
    \caption{\textbf{Zero-shot results for text-based satellite image retrieval}. The columns show the top 4 images returned given the text query on top. A model can be queried using general landscape descriptions (\eg `desert', `ocean', `forest'). In addition, specific wildlife text such as `ibex' and `house finch' can be used as queries to view the types of environment they inhabit. The `ibex' inhabits mountains and high elevation areas, `american alligators' are found in swamps and the coast, the `house finch' is commonly found in urban areas---consistent with the retrieved satellite images shown. More examples are available in \cref{fig:zero-shot-results-addtl} (Appendix).}
    \label{fig:zero-shot-results}
    \vspace{-5mm}
\end{figure}

\begin{table*}[!t]
    \setlength{\tabcolsep}{4pt}
    \small
    \begin{center}
    \resizebox{0.93\linewidth}{!}{
    \begin{tabular}{c c c c | c c  | c c  | c c  | c c  | c }
    \toprule
     &  &  &  & \multicolumn{2}{c}{UCM \cite{yang2010bag_ucmdataset} } & \multicolumn{2}{c}{AID \cite{xia2017aid}} & \multicolumn{2}{c}{RESISC45 \cite{cheng2017remote_resisc45dataset}}  & \multicolumn{2}{c|}{So2Sat20k \cite{zhu2020so2sat}} &  \\
    loc & env & text & img-a & Random & ImageNet & Random & ImageNet & Random & ImageNet & Random & ImageNet & Average \\
    &  &  &  & ResNet50 & ViT-B/16 & ResNet50 & ViT-B/16 & ResNet50 & ViT-B/16 & ResNet50 & ViT-B/16 & \\
    \midrule
    \midrule
     &  &  &  & 24.3\% & 93.2\% & 25.2\% & 84.4\% & 25.5\% & 88.2\% & 5.9\% & 41.8\% & 48.6\% \\
    \checkmark &  &  &  & 44.2\% & 95.0\% & 41.6\% & 85.1\% & 43.0\% & 89.3\% & 18.3\% & 43.4\% & 57.5\% \\
    \checkmark & \checkmark &  &  & 60.0\% & 95.0\% & 48.7\% & 86.2\% & 48.2\% & 88.8\% & 25.2\% & 44.2\% & 62.0\% \\
    \checkmark & \checkmark & \checkmark &  & 70.0\% & 95.4\% & 55.6\% & 86.2\% & 58.9\% & 89.7\% & 27.9\% & 45.0\% & 66.1\% \\
    \checkmark & \checkmark & \checkmark & \checkmark & \textbf{79.9\%} & \textbf{97.2\%} & \textbf{68.2\%} & \textbf{88.9\%} & \textbf{74.7\%} & \textbf{93.0\%} & \textbf{42.3\%} & \textbf{55.2\%} & \textbf{74.9\%} \\
    \bottomrule
    \end{tabular}
    }
    \vspace{-5pt}
    \caption{\textbf{Ablation of various components of WildSAT.} The best performance is a combination of all components: location (loc), environmental covariates (env), text (text), and satellite image augmentations (img-a). Results are presented for a randomly initialized ResNet50 model and an ImageNet pre-trained ViT-B/16 model with the numbers referring to linear probing accuracy.}
    \label{table:summary-ablation}
    \end{center}
    \vspace{-8mm}
\end{table*}

\subsection{Bird Species Encounter Rate Prediction}
\cref{table:satbird-results} shows WildSAT complementing existing methods on bird species encounter rate prediction. We take models SatlasNet (Swin-B)~\cite{bastani2023satlaspretrain} and SatMAE (ViT-L/16)~\cite{cong2022satmae} presented in SatBird~\cite{teng2024satbird} that use satellite images to predict encounter rates of all bird species in an area. TaxaBind~\cite{taxabind2025} is also added as a baseline. Using WildSAT improves these existing methods. In addition, since most of the data used in WildSAT are from the USA and Europe, we see more performance improvement in USA bird encounter rates.

\subsection{Zero-shot Image Retrieval} 
When trained using our WildSAT framework, we observe that models  learn wildlife-specific attributes.
By using the frozen satellite image encoder and a large language model, a user can input text to query satellite images.
The top $k$ images with the most similar embeddings to the text embeddings (computed using cosine similarity) can be retrieved (\cref{fig:architecture}b.2).
\cref{fig:zero-shot-results} displays examples of satellite images retrieved given different text queries.
General descriptions of landscapes or locations can be used for querying such as `desert', `ocean', `forest', or `grassland'. 
At the same time, specific wildlife text can also be used as queries such as `ibex' or `gull'. We see that zero-shot retrieval returns images of the habitat of the corresponding wildlife.
This feature could potentially be used to find habitats for species with limited observation data, and can serve as a reference for species distribution studies and biodiversity monitoring.

\subsection{Ablations}
Our WildSAT framework is composed of multiple components: a satellite image encoder, location encoder, and text encoder.
We investigate the contribution of each of these components in an ablation study.
In \cref{table:summary-ablation} we ablate two models: a randomly initialized ResNet50 and an ImageNet pre-trained ViT-B/16. 
Columns with check marks on `loc' and/or `env' indicate the use of the location encoder ($\mathcal{L}_{\text{loc}}$ term in Eqn.~\ref{eq:optimization}). 
The location encoder can use either only location as input or use both location and environmental covariates. 
The `text' and `img-a' columns indicate the use of the large language model ($\mathcal{L}_{\text{txt}}$) and contrastive loss on the augmented satellite images ($\mathcal{L}_{\text{img}}$), respectively. A complete ablation of all combinations of WildSAT components for other models is also provided in \cref{table:full-ablation}-\ref{table:peft-results} (Appendix).

\noindent\textbf{Each modality enhances performance.} Larger improvements are observed for the randomly initialized model, given its lower starting accuracy and greater potential for improvement (\cref{table:summary-ablation}).
Nonetheless, similar improvement trends are seen with both types of models.
Improvements from location and environmental covariates can be from associating satellite images with particular locations and corresponding environmental characteristics (\ie satellite images are mapped to specific latitudes and longitudes that are along the coast, or in the mountains).
The addition of text further improves the performance while enabling zero-shot image retrieval capabilities.
Text provides a rich source of information with more detailed descriptions of areas.
Adding the satellite image loss term likely improves general image understanding, such as the model learning the rotation invariance of satellite images.
Overall, each component of WildSAT contributes to performance improvements.

\noindent\textbf{Different modalities can strengthen models by covering model-specific gaps.} 
We hypothesize that models trained on satellite imagery datasets benefit primarily from location and text supervision associated with wildlife observation.
This hypothesis is supported by observations that self-supervised models such as SeCo, SatMAE, and Prithvi---trained with objectives similar to WildSAT’s image self-supervision term on satellite datasets---still achieve significant gains (\cref{fig:linear-probing}). 
Similarly, models like SatlasNet, which are trained with large-scale supervised learning on satellite images, also benefit. On the other hand, ImageNet pre-trained models benefit from the additional satellite image supervision. These results highlight the complementary nature of WildSAT’s supervision compared to existing datasets, which primarily focus on anthropogenic labels. \cref{table:full-ablation} (Appendix) further supports this finding.

\noindent\textbf{PEFT preserves out-of-domain pre-training representations.} Applying PEFT to out-of-domain pre-trained models (\eg ImageNet, CLIP) helps preserve their original representations, preventing performance degradation---a crucial advantage for large models like ViT, where fine-tuning all weights risks shifting parameters suboptimally (\cref{table:clip-model-results-summary}). In contrast, in-domain pre-trained models (\eg SeCo) benefit more from full fine-tuning, as WildSAT's similar data leads to a non-disruptive shift as shown in \cref{table:peft-results} (Appendix). 

\noindent\textbf{SINR location encoder achieves the best performance.} We explore different location encoders in \cref{table:location-encoder-ablation} (Appendix). Comparing no location encoder (\ie using position encoded latitude and longitude), SINR, and SatCLIP, SINR delivers best overall performance, likely due to its more informative location representations that better incorporate species habitat information. 

\subsection{Limitations}
While we show that WildSAT improves satellite image representations and has promising zero-shot performance, the datasets we train on contain inherent limitations that could affect model use.
The US and Europe are overrepresented in the data as most citizen contributed data currently come from these locations.
Underrepresented areas in Asia, Africa, Australia, and South America are likely to see less accurate results especially on satellite image retrieval.
LLMs are further prone to generating hallucinations, which could impact model output reliability.

%
%
%
%
\section{Conclusion}
\label{sec:conclusion}

While satellite images are often used to interpolate sparse wildlife observations to create species range maps, our work demonstrates that these observations also provide a rich source of supervision for learning satellite image representations. WildSAT can not only learn high-quality representations from scratch but also improve performance of strong pre-trained models, such as those trained on ImageNet and satellite imagery datasets, across a range of satellite imagery tasks. We attribute this success to the global scale of community efforts, which document diverse wildlife observations through platforms like iNaturalist and eBird and record detailed species attributes on sources like Wikipedia. This supervisory signal complements existing satellite datasets, which often focus on anthropogenic labels, by introducing a broader ecological perspective. As these resources continue to expand in both observational scale (\eg geographical and taxonomic scope) and modality (\eg incorporating sound, aerial imagery), they offer even greater potential for improving WildSAT’s representations.

\paragraph{Acknowledgements.}
We thank the iNaturalist community for providing the data used for training.
Experiments were performed on the University of Massachusetts GPU cluster funded by the Mass.~Technology Collaborative. RD and SM were supported in part by NASA grant 80NSSC22K1487 and NSF grant 2329927. 

{
    \small
    \bibliographystyle{ieeenat_fullname}
    \bibliography{main}
}


\appendix 

\maketitlesupplementary

\setcounter{table}{0}
\renewcommand{\thetable}{A\arabic{table}}
\setcounter{figure}{0}
\renewcommand{\thefigure}{A\arabic{figure}}

\input{suppl}

\end{document}

%% file: suppl.tex
\section{Datasets}
\subsection{Training Data Distribution}

\begin{figure}[h]
    \begin{center}
    \includegraphics[scale=0.45]{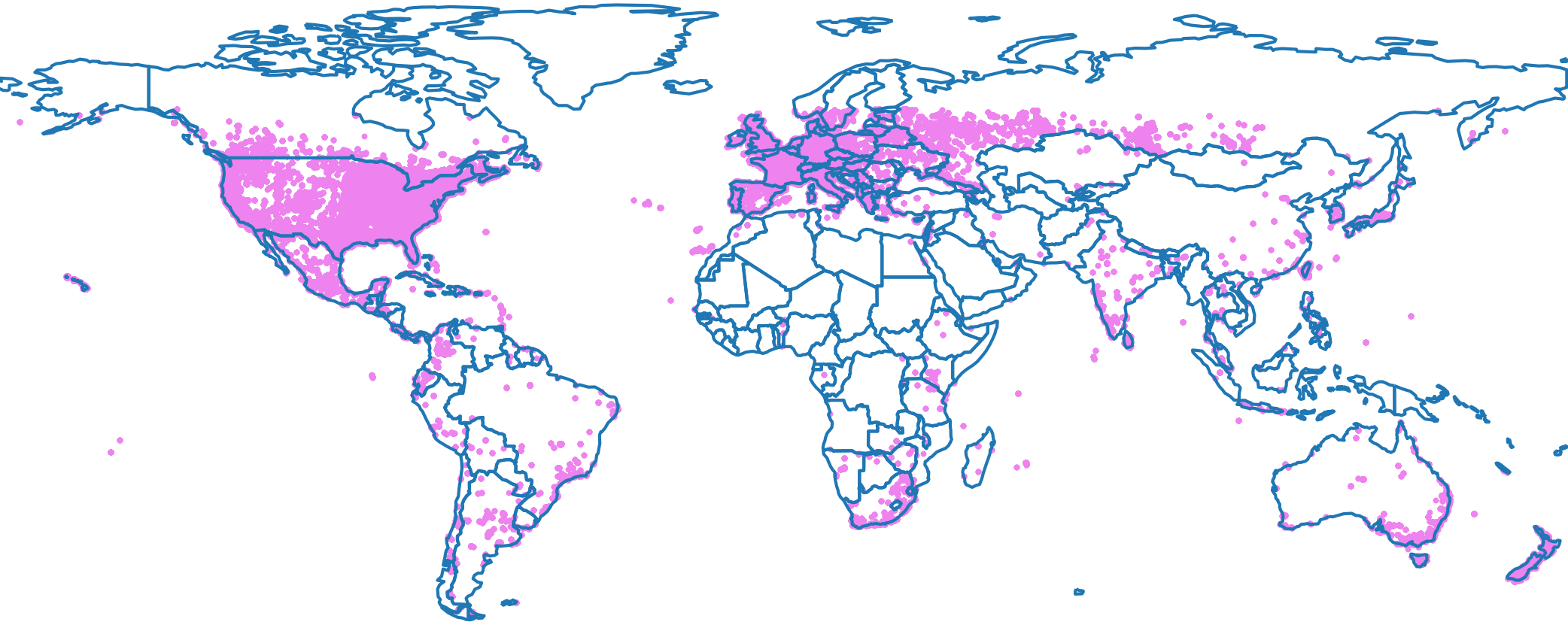}
    \end{center}
    \vspace{-3mm}
    \caption{\textbf{Distribution of data points with satellite image, environmental covariates, and text}. The alignment of different modalities is guided by the geographic distribution of species.}
    \label{fig:data-distribution}
    \vspace{-5mm}
\end{figure}

\cref{fig:data-distribution} shows the spatial distribution of all data we collected across the globe. Most of the data are from United States and Europe, corresponding to the wildlife observation data available from citizen science platforms~\cite{inatWeb, eBirdWeb}.
Despite this, we show that models trained with WildSAT can still generalize to areas beyond the US and Europe, with segmentation improvements even in areas like Africa (see~\cref{fig:segmentation}).

\subsection{Satellite Image Evaluation Datasets}
Below we briefly describe the different satellite image classification datasets used for evaluation. For each of the datasets, the splits for training, validation, and testing follow those provided from the respective sources.

\noindent\textbf{UCM}~\cite{yang2010bag_ucmdataset} is an image classification dataset that contains 21 classes with 100 each covering USA. Each image is 256$\times$256 with a resolution of 1~ft. 

\noindent\textbf{AID}~\cite{xia2017aid} is an image classification dataset that contains 30 class,  each containing from 220 to 400 images from Google Earth. Each image is 600$\times$600 with a resolution ranging from 0.5 to 8~m.

\noindent\textbf{RESISC45}~\cite{cheng2017remote_resisc45dataset} is an image classification dataset that contains 45 classes with 700 images each, sourced from Google Earth. An image is 256$\times$256 with a resolution ranging from 0.2 to 30~m.

\noindent\textbf{FMoW}~\cite{christie2018functional_fmow} is an image classification dataset that contains 63 classes from over 200 countries with a total of over 400k images from QuickBird, GeoEye, and WorldView satellites. Images vary in size and resolution and each class has a different number of images.

\noindent\textbf{EuroSAT}~\cite{helber2019eurosat} is an image classification dataset that contains 10 classes of land use and land cover from Europe. Each image is 64$\times$64 with 10~m resolution. The dataset has 27k images with each class having a different number of images.

\noindent\textbf{So2Sat20k}~\cite{zhu2020so2sat} is an image classification dataset that contains 17 classes across different climate zones with global coverage. The full dataset contains 400k pairs of Sentinel-1 and Sentinel-2 images. We use the GEO-Bench~\cite{lacoste2024geobench} version referred to as ``So2Sat20k" which contains 20k training samples.

\noindent\textbf{BigEarthNet20k} (BEN20k)~\cite{sumbul2019bigearthnet} is a multi-label classification dataset with 43 classes. The full dataset is from 10 countries in Europe with 590k Sentinel-2 images. We use the GEO-Bench~\cite{lacoste2024geobench} version ``BEN20k" which contains 20k training samples.

\noindent\textbf{Cashew1k}~\cite{yin2023mapping} is a segmentation dataset with 7 categories mapping cashew plantations in Benin, Africa. It contains more than 1k training examples of Sentinel-2 images. We use the available dataset in GEO-bench.

\noindent\textbf{SACrop3k}~\cite{sacrop} is a segmentation dataset with 10 categories mapping crop types in South Africa. It contains 3k training samples of Sentinel-2 images. We use the dataset provided in GEO-bench.

\section{Additional Implementation Details}
\label{sec:addtl-implementation-details}
\paragraph{WildSAT Training Details.} 
We train each base model on the WildSAT framework for 25 epochs using an Adam optimizer with a learning rate of $1\times 10^{-4}$, with an embedding dimension $d=512$, and a batch size of 64.
Random cropping, resizing, jitter, and channel mixing are applied on satellite images as augmentations.
Each satellite image is paired with a wildlife observation location for positive samples.
For each of these pairs, a section of text is randomly sampled from the Wikipedia~\cite{wikipedia} page of the species.
A satellite image of the same location, but from a different time, is also randomly sampled for image augmentation.
Negative samples are randomly selected from other species and locations that do not correspond to the observed wildlife locations. 
For multiple species that are found in the same location, multiple texts are associated with the same location.
For the same species present in different locations, the text for each location is randomly sampled from the same Wikipedia page (\ie each location would correspond to a random section on the same Wikipedia page). 
Downstream tasks follow the train/val/test split provided in each of the datasets. Training takes an average of 8 hours on 2 NVIDIA L40S.

\paragraph{Satellite Image Filtering.} 
All satellite images are from Sentinel-2A and Sentinel-2B. 
We follow the same data collection procedure from SatlasPretrain~\cite{bastani2023satlaspretrain}, where satellite images are downloaded from EU's Sentinel Data~\cite{esa2022sentinel}.
Each image is 512$\times$512 pixels with a 10~m resolution per pixel. 
Only images that are tagged with significantly less cloud cover from~\cite{bastani2023satlaspretrain} are used.
In addition, we only use satellite images that were taken in the same time range as the wildlife observation data (from 2017 to 2021).
This is done since we do not use the exact observation date and time as an input to the model; we consider all observations throughout the time range. At the same time, the text descriptions we use also refer to all types of habitats regardless of time of year.

\paragraph{Multi-spectral Baselines.} Prithvi-100M~\cite{jakubik2310foundation_prithvi} and SatCLIP~\cite{klemmer2023satclip} originally use multi-spectral data in their pre-training. However, for general applicability and easy comparisons with other models, we only use RGB bands in Table 1. When WildSAT is applied to these models, we only fine-tune with the three bands, and set other bands to zero. At the same time, when applying both the base models and WildSAT fine-tuned models on downstream satellite image datasets, we also set other bands to zero. We also explore using multiple bands as inputs in \cref{table:multispectral-results}, and discuss the results in the next section. In this case, when a band used by a model is not available in the dataset, we set the values of the band to zero. Otherwise, we use all the bands available.

\section{Additional Results}

\begin{table*}[!t]
    \setlength{\tabcolsep}{4pt}
    \small
    \begin{center}
    \begin{tabular}{l l|r r|r r|r r|r r|r r|r r|r r}
    \toprule
    \multirow{3}{*}{} & \multirow{3}{*}{\textbf{Encoder}} & \multicolumn{2}{c}{UCM  } & \multicolumn{2}{c}{AID } & \multicolumn{2}{c}{RESISC45 } & \multicolumn{2}{c}{FMoW } & \multicolumn{2}{c}{EuroSAT } & \multicolumn{2}{c}{So2Sat20k }  & \multicolumn{2}{c}{BEN20k } \\
     & & \multicolumn{2}{c}{\cite{yang2010bag_ucmdataset} } & \multicolumn{2}{c}{\cite{xia2017aid}} & \multicolumn{2}{c}{\cite{cheng2017remote_resisc45dataset}} & \multicolumn{2}{c}{\cite{christie2018functional_fmow}} & \multicolumn{2}{c}{\cite{helber2019eurosat}} & \multicolumn{2}{c}{\cite{zhu2020so2sat}}  & \multicolumn{2}{c}{\cite{sumbul2019bigearthnet}} \\\cline{3-16}
     & & Base & +WS & Base & +WS & Base & +WS & Base & +WS & Base & +WS & Base & +WS & Base & +WS \\
    \midrule
    \midrule
     & ImageNet~\cite{deng2009imagenet} & 93.2 & 97.5 & 84.4 & 88.9 & 88.2 & 93.0 & 43.8 & 51.4 & 94.5 & 97.3 & 41.8 & 55.2 & 52.3 & 58.2 \\
     & MoCov3~\cite{chen2021empirical_mocov3} & 94.2 & 95.1 & 86.0 & 86.9 & 89.1 & 90.3 & 51.1 & 52.9 & 95.9 & 97.1 & 47.6 & 56.6 & 51.6 & 57.0 \\
     & CLIP~\cite{radford2021learning_clip} & 94.5 & 96.3 & 86.3 & 88.0 & 92.1 & 93.0 & 51.5 & 52.8	& 92.2 & 97.1 & 37.6 & 49.7 & 47.1 & 59.1 \\
     & Prithvi-100M*~\cite{jakubik2310foundation_prithvi} & 49.7 & 85.5 & 35.9 & 71.2 & 42.6 & 73.5 & 19.2 & 30.5 & 67.3 & 93.5 & 21.5 & 45.1 & 33.6 & 50.6 \\
     \rot{\rlap{~ViT-B/16}}
    
     & SatCLIP*~\cite{klemmer2023satclip} & 38.2 & 50.3 & 37.4 & 46.4 & 40.4 & 46.2 & 19.0 & 20.1 & 74.6 & 79.4 & 39.0 & 43.1 & 27.0 & 28.7 \\
     & Random weights & 4.1 & 75.5 & 3.8 & 62.1 & 1.9 & 62.4 & 8.0 & 26.0 & 11.1 & 90.4 & 5.9 & 46.8 & 0.0 & 51.2 \\
    
    \midrule
     & ImageNet~\cite{deng2009imagenet} & 94.0 & 96.9 & 87.9 & 89.0 & 90.4 & 91.8 & 47.6 & 50.7 & 96.2 & 97.3 & 48.3 & 51.5 & 54.1 & 57.7 \\
     & SatlasNet~\cite{bastani2023satlaspretrain} & 89.6 & 91.2 & 74.3 & 81.2 & 80.2 & 86.5 & 31.8 & 44.6 & 90.8 & 95.5 & 36.4 & 53.1 & 48.7 & 56.5 \\
    \rot{\rlap{~Swin-T}}
     & Random weights & 21.0 & 81.7 & 19.5 & 72.0 & 19.9 & 74.9 & 12.1 & 33.4 & 59.9 & 92.7 & 21.9 & 45.9 & 9.8 & 52.4 \\
    \midrule
     & ImageNet~\cite{deng2009imagenet} & 94.2 & 93.6 & 87.8 & 86.7 & 90.5 & 90.1 & 47.3 & 46.0 & 95.5 & 96.0 & 36.1 & 46.6 & 55.8 & 57.5 \\
     & MoCov3~\cite{chen2021empirical_mocov3} & 92.0 & 93.5 & 83.0 & 83.3 & 88.0 & 87.6 & 50.2 & 45.7 & 93.5 & 95.1 & 27.2 & 42.5 & 46.6 & 53.8 \\
     & SatlasNet~\cite{bastani2023satlaspretrain} & 86.8 & 90.1 & 72.5 & 79.4 & 81.8 & 85.4 & 34.7 & 42.4 & 93.5 & 95.4 & 33.9 & 44.8 & 44.9 & 56.4 \\
     & SeCo~\cite{manas2021seasonal_seco} & 86.1 & 88.8 & 74.3 & 79.6 & 80.2 & 86.3 & 35.9 & 42.8 & 89.7 & 95.5 & 39.9 & 46.0 & 44.3 & 57.3 \\
    \rot{\rlap{~ResNet50}}
     & SatCLIP*~\cite{klemmer2023satclip} & 69.4 & 76.2 & 63.1 & 71.8 & 70.2 & 78.8 & 36.2 & 39.9 & 83.4 & 92.9 & 45.4 & 44.9 & 42.3 & 48.2 \\
     & Random weights & 24.7 & 79.9 & 22.3 & 68.2 & 24.5 & 74.7 & 12.7 & 36.9 & 65.2 & 92.2 & 5.9 & 42.3 & 19.9 & 51.3 \\
    \midrule
    \midrule
    \multicolumn{2}{l}{Overall average} & 68.8 & \textbf{86.1} & 61.2 & \textbf{77.0} & 65.3 & \textbf{81.0} & 33.4 & \textbf{41.1} & 80.2 & \textbf{93.8} & 32.6 & \textbf{47.6} & 38.5 & \textbf{53.1} \\
    \multicolumn{2}{l}{Average w/o random} & 81.8 & \textbf{87.9} & 72.7 & \textbf{79.4} & 77.8 & \textbf{83.5} & 39.0 & \textbf{43.3} & 88.9 & \textbf{94.3} & 37.9 & \textbf{48.3} & 45.7 & \textbf{53.4} \\
    \bottomrule
    \end{tabular}
    \vspace{-6pt}
    \caption{\textbf{Results of linear probing different models on seven downstream datasets without (Base) and with (+WS) WildSAT fine-tuning}. Accuracy is reported for all datasets except BEN20k that reports micro F1 score. Base refers to the original models specified as the encoder and +WS refers to the same models further trained on the species observation data. Fine-tuning models with species observation show significant improvement over the base models. Both Prithvi-100M and SatCLIP are pre-trained with multispectral images, but for consistency across downstream datasets and models, only RGB bands are used. We include results on multispectral images in \cref{table:multispectral-results}.}
    \label{table:overall-results}
    \end{center}
    \vspace{-9mm}
\end{table*}

\begin{table*}[!t]
    \setlength{\tabcolsep}{3pt}
    \small
    \begin{center}
    \begin{tabular}{l l|r r|r r|r r|r r|r r|r r|r r}
    \toprule
    \multirow{3}{*}{} & \multirow{3}{*}{\textbf{Encoder}} & \multicolumn{2}{c}{UCM  } & \multicolumn{2}{c}{AID } & \multicolumn{2}{c}{RESISC45 } & \multicolumn{2}{c}{FMoW } & \multicolumn{2}{c}{EuroSAT } & \multicolumn{2}{c}{So2Sat20k }  & \multicolumn{2}{c}{BEN20k } \\
     & & \multicolumn{2}{c}{\cite{yang2010bag_ucmdataset} } & \multicolumn{2}{c}{\cite{xia2017aid}} & \multicolumn{2}{c}{\cite{cheng2017remote_resisc45dataset}} & \multicolumn{2}{c}{\cite{christie2018functional_fmow}} & \multicolumn{2}{c}{\cite{helber2019eurosat}} & \multicolumn{2}{c}{\cite{zhu2020so2sat}}  & \multicolumn{2}{c}{\cite{sumbul2019bigearthnet}} \\\cline{3-16}
     & & Base & +WS & Base & +WS & Base & +WS & Base & +WS & Base & +WS & Base & +WS & Base & +WS \\
    \midrule
    \midrule
    ResNet50 & ImageNet1K V2~\cite{recht2019imagenet} & 94.2 & 93.6 & 87.8 & 86.7 & 90.5 & 90.1 & 47.3 & 46.0 & 95.5 & 96.0 & 36.1 & 46.6 & 55.8 & 57.5 \\
    ResNet50 & ImageNet1K V1~\cite{deng2009imagenet} & 92.5 & 93.5 & 90.4 & 88.8 & 85.1 & 84.7 & 40.7 & 37.0 & 88.0 & 94.9 & 38.8 & 48.2 & 46.7 & 53.7 \\
    ViT-L/16 & SatMAE~\cite{cong2022satmae} & 23.8 & 86.1 & 25.1 & 70.6 & 26.1 & 74.6 & 13.9 & 33.4 & 48.3 & 94.5 & 15.6 & 48.6 & 18.4 & 51.0 \\
    \bottomrule
    \end{tabular}
    \vspace{-6pt}
    \caption{\textbf{Additional linear probing results on satellite image classification datasets}. Accuracy is reported for all datasets except BEN20k that reports micro F1 score. `Base' refers to the original models specified as the encoder and `+WS' refers to the same models further trained with WildSAT. Consistent with results thus far, fine-tuning models with species observation generally show significant improvement over the base models.}
    \label{table:addtl-classification-results}
    \end{center}
    \vspace{-6mm}
\end{table*}

\begin{table*}[!t]
    \small
    \begin{center}
    \begin{tabular}{l | c c c c c c c }
    \toprule
    \multirow{2}{*}{} & UCM & AID & RESISC45 & FMoW & EuroSAT & So2Sat20k  & BEN20k \\
     & \cite{yang2010bag_ucmdataset} & \cite{xia2017aid} & \cite{cheng2017remote_resisc45dataset} & \cite{christie2018functional_fmow} & \cite{helber2019eurosat} & \cite{zhu2020so2sat}  & \cite{sumbul2019bigearthnet} \\
    \midrule
    
    TaxaBind~\cite{taxabind2025} & 80.5 & 67.7 & 72.6 & 31.2 & 85.2 & 33.9 & 47.6 \\
    GRAFT~\cite{mall2024remote_graft} & 81.1 & 76.1 & 83.3 & 39.3 & 90.9 & 36.6 & 48.0 \\
    RemoteCLIP~\cite{liu2024remoteclip} & 96.1 & 86.1 &	90.9 &	45.7 &	93.3 &	35.5 &	49.4 \\
    CLIP~\cite{radford2021learning_clip} & 94.5 & 86.3 & 92.1 & 51.5 & 92.2 & 37.6 & 47.1 \\
    \midrule
    WildSAT (Ours) & \textbf{96.3} & \textbf{88.0} & \textbf{93.0} & \textbf{52.8} & \textbf{97.1} & \textbf{49.7} & \textbf{59.1} \\

    \bottomrule
    \end{tabular}
    \vspace{-6pt}
    \caption{\textbf{Linear probing results on downstream satellite classification datasets  using models with CLIP as the base.} Results are reported as accuracy, except for  BEN20k which uses micro F1. TaxaBind, GRAFT, and RemoteCLIP fine-tune a CLIP backbone and use additional modalities such as text, ground images, and satellite images for cross-modal tasks. WildSAT outperforms both the standard CLIP and the previous methods that also fine-tune on CLIP.}
    \label{table:clip-model-results}
    \end{center}
    \vspace{-8mm}
\end{table*}

\begin{table*}[!t]
	\setlength{\tabcolsep}{3pt}
	\small
	\begin{center}
	\begin{tabular}{l l|r r|r r}
	\toprule
	\multirow{3}{*}{} & \multirow{3}{*}{\textbf{Encoder}} & \multicolumn{2}{c}{So2Sat20k }  & \multicolumn{2}{c}{BEN20k } \\
		& & \multicolumn{2}{c}{\cite{zhu2020so2sat}}  & \multicolumn{2}{c}{\cite{sumbul2019bigearthnet}} \\\cline{3-6}
		& & Base & +WS & Base & +WS \\
	\midrule
	\midrule
	ViT-B/16 & Prithvi-100M~\cite{jakubik2310foundation_prithvi} & 28.7 & 50.1 & 34.4 & 53.9 \\
	ViT-B/16 & SatCLIP~\cite{klemmer2023satclip} & 48.8 & 48.8 & 33.4 & 36.1 \\
	ResNet50 & SatCLIP~\cite{klemmer2023satclip} & 45.8 & 46.3 & 46.7 & 54.3 \\
    \midrule
	 Average  &  & 41.1 & \textbf{48.4} & 38.2 & \textbf{48.1} \\
	\bottomrule
	\end{tabular}
	\vspace{-6pt}
	\caption{\textbf{Linear probing results on models using multispectral images}. Accuracy is reported for So2Sat20k and micro F1 score for BEN20k. `Base' refers to the original models specified as the encoder and `+WS' refers to the same models further trained with WildSAT. We see improved base model performance with using more bands (compared to using only RGB), and show that the addition of WildSAT further improves performance even with multispectral models and datasets.}
	\label{table:multispectral-results}
	\end{center}
	\vspace{-6mm}
\end{table*}

\subsection{Satellite Image Classification}
\paragraph{Linear Probing results.} \cref{table:overall-results} shows the raw numbers from which \cref{fig:linear-probing} was generated.

\paragraph{ImageNet V1 results.} \cref{table:addtl-classification-results} shows results on the ImageNet V1 base model that previous works have used~\cite{manas2021seasonal_seco,mall2023change_caco}.
The results in Table~1 in the main paper include a base model using ImageNet V2 (also included in \cref{table:addtl-classification-results} for reference) which has generally better performance across the downstream satellite image datasets.
\cref{table:addtl-classification-results} additionally shows results on SatMAE~\cite{cong2022satmae}, a ViT-L/16 model that was pre-trained with the MAE framework on satellite images.
Similar to previous results, WildSAT improves performance across the seven satellite image classification datasets evaluated.

\paragraph{Multi-spectral results.} \cref{table:multispectral-results} shows additional results when using multispectral images as input to the models that support this capability. Prithvi-100M and SatCLIP accept multiple bands as input, while So2Sat20k and BEN20k are datasets on GEO-bench that contain multiple bands. We find that WildSAT also improves on these multispectral models, demonstrating the broad applicability of our method.

\paragraph{WildSAT outperforms CLIP-based models.} 
\cref{table:clip-model-results} displays the result for each evaluation dataset across different CLIP-based models.
All models in the table starts with a pre-trained CLIP ViT-B/16 model~\cite{radford2021learning_clip}. TaxaBind\cite{taxabind2025} and GRAFT~\cite{mall2024remote_graft} use additional modalities such as ground images, text, and audio to improve model performance on cross-modal tasks such as zero-shot image-text retrieval.
However, we show that while these same models do well on zero-shot tasks, they tend to ``forget" some of the original image representations, with linear probing performance on downstream datasets lower than that of the original CLIP model.
With WildSAT applied to CLIP, we show that we can outperform not only other CLIP-based models, \ie GRAFT and TaxaBind, but also outperform the standard CLIP model across all the datasets in the evaluation.
We hypothesize we can prevent ``forgetting" by applying parameter efficient fine-tuning on out-of-domain pre-trained models such as CLIP. We further show this in \cref{table:peft-results}.

\begin{figure*}[!t]
    \begin{center}
    \includegraphics[scale=0.51]{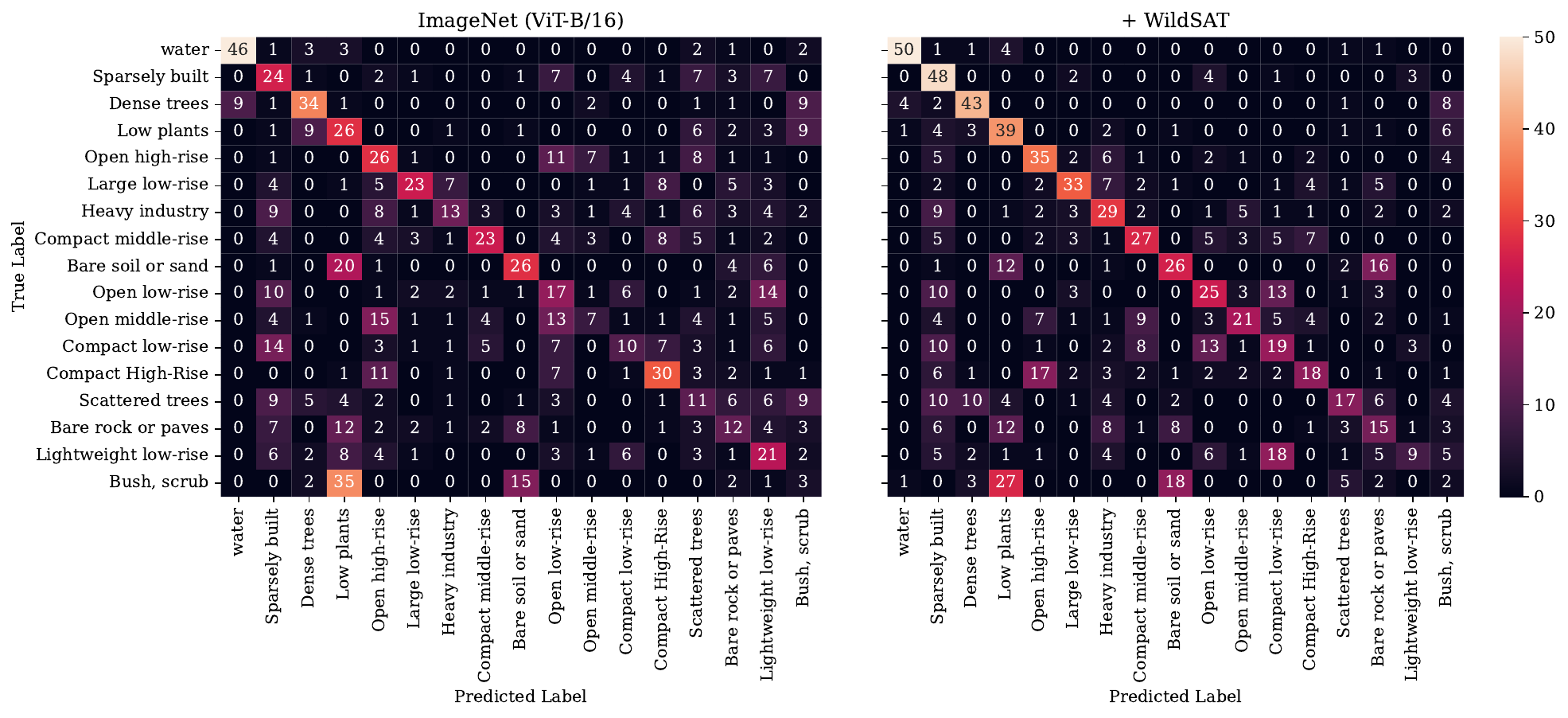}
    \end{center}
    \vspace{-10pt}
    \caption{\textbf{Confusion matrices comparing the predictions of an ImageNet base model and a WildSAT fine-tuned model on the So2Sat20k dataset~\cite{zhu2020so2sat}}. Both models use a ViT-B/16 architecture. Each matrix displays the result on the provided So2Sat20k test set in GEO-Bench~\cite{lacoste2024geobench}.
    }
    \label{fig:confusion-matrix-so2sat}
    \vspace{-6mm}
\end{figure*}

\begin{table*}[!t]
    \small
    \begin{center}
    \begin{tabular}{l |r r r | r r r }
    \toprule	
	& \multicolumn{3}{c|}{EuroSAT~\cite{helber2019eurosat}} & \multicolumn{3}{c}{So2Sat20k~\cite{zhu2020so2sat}		} \\
	& AnnualCrop & SeaLake & Highway & Water & Low plants & Heavy industry \\
	\midrule
	TaxaBind~\cite{taxabind2025} & 1.3 & 20.4 & 3.7 & 0.0 & 0.0 & 0.0 \\
	GRAFT~\cite{mall2024remote_graft} & 33.3 & 72.1 & \textbf{55.6} & 16.4 & 0.0 & \textbf{13.4} \\
        RemoteCLIP~\cite{liu2024remoteclip} & 19.6 & 15.9 & 5.3 & 36.5 & 5.9 & 0.0 \\
	\midrule
	WildSAT (Ours) & \textbf{46.5} & \textbf{87.4} & 0.4 & \textbf{60.9} & \textbf{10.4} & 0.0 \\
	\bottomrule
    \end{tabular}
    \vspace{-3mm}
    \caption{\textbf{Zero-shot image classification F1 score on different classes of EuroSAT and So2Sat20k with CLIP-based models}. Since WildSAT is geared towards habitat-related classes, the coverage of zero-shot classification is less effective beyond natural concepts. WildSAT does well on natural classes like `SeaLake' and `Water', but struggles on anthropogenic classes like `Highway' and `Heavy industry'.}
    \label{table:quant-zero-shot}
    \end{center}
    \vspace{-6mm}
\end{table*}

\begin{table*}[!t]
    \small
    \begin{center}
    \begin{tabular}{l |r r r r | r r r r }
    \toprule	

	& \multicolumn{4}{c|}{Cashew1k~\cite{yin2023mapping}} & \multicolumn{4}{c}{SAcrop3k~\cite{sacrop}} \\
	& \multicolumn{2}{c}{Accuracy}  & \multicolumn{2}{c|}{IoU} &  \multicolumn{2}{c}{Accuracy}  & \multicolumn{2}{c}{IoU}  \\
        \midrule
	& Base & +WS & Base & +WS & Base & +WS & Base & +WS \\
        \midrule
	ImageNet~\cite{deng2009imagenet} & 91.6\% & 91.9\% & 70.3\% & 70.6\% & 60.7\% & 62.3\% & 24.3\% & 25.0\% \\
	MoCov3~\cite{chen2021empirical_mocov3} & 92.4\% & 93.2\% & 71.4\% & 73.3\% & 60.7\% & 60.8\% & 22.9\% & 24.9\% \\
	SeCo~\cite{manas2021seasonal_seco} & 86.7\% & 93.2\% & 62.6\% & 73.3\% & 59.2\% & 59.4\% & 22.3\% & 22.8\% \\
	SatlasNet~\cite{bastani2023satlaspretrain} & 82.5\% & 91.8\% & 55.2\% & 71.0\% & 56.1\% & 57.1\% & 19.4\% & 20.5\% \\
	Random & 69.8\% & 92.9\% & 40.1\% & 72.6\% & 54.7\% & 56.3\% & 18.0\% & 20.3\% \\
	\midrule
	& 84.6\% & \textbf{92.6\%} & 59.9\% & \textbf{72.2\%} & 58.3\% & \textbf{59.2\%} & 21.4\% & \textbf{22.7\%} \\
     \bottomrule
     \end{tabular}
     \vspace{-3mm}
     \caption{\textbf{Downstream satellite image segmentation results}. WildSAT can also improve on satellite segmentation tasks across different models and datasets. All models use the frozen pre-trained encoders with a convolutional-based decoder trained for each of the downstream tasks. `Base' refers to the original models specified as the encoder and `+WS' refers to the same models further trained on the species observation data.}
     \label{table:segmentation-results}
     \end{center}
     \vspace{-6mm}
\end{table*}

\begin{figure*}[!t]
    \begin{center}
    \includegraphics[scale=0.43]{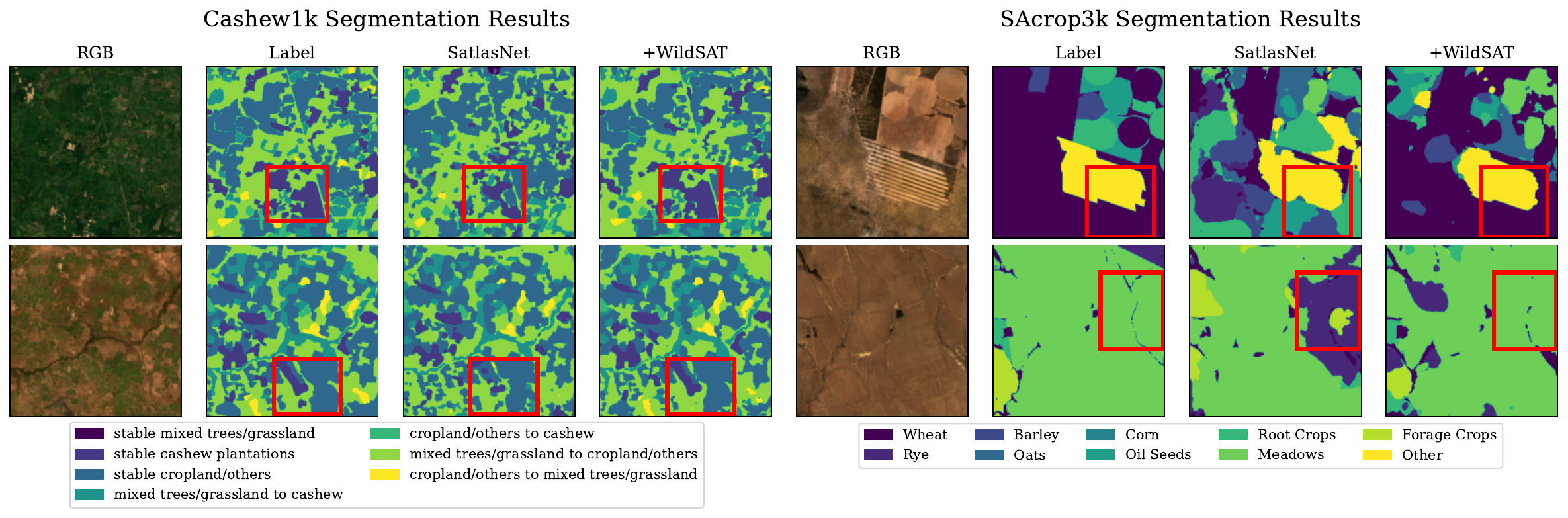}
    \end{center}
    \vspace{-16pt}
    \caption{\textbf{Visualization of segmentation results using SatlasNet without and with WildSAT}. WildSAT can more accurately identify classes `stable cashew plantations' and `stable cropland/others' in Cashew1k~\cite{yin2023mapping}, and improve on the identification of `Rye' and `Meadows' in SAcrop3k~\cite{sacrop}. Some areas of improvement are highlighted in red boxes.}
    \label{fig:segmentation}
    \vspace{-6mm}
\end{figure*}

\paragraph{WildSAT reduces errors on habitat-related classes.} \cref{fig:confusion-matrix-so2sat} shows the confusion matrix of a sample result on the So2Sat20k test set. 
The matrix on the left shows the result of a base ImageNet pre-trained model, and the right matrix shows the result when WildSAT is applied.
We show that WildSAT improvements on the true positive counts along the diagonal are largely due to fewer false positives on habitat-related classes.
Looking at the second row of both matrices (class `Sparsely built'), the true positive count doubled from 24 to 48. A lot of this improvement comes from less false positives on `Scattered trees' (from 7 false positives to 0), `Bare rock or paves' (from 3 false positives to 0), and `Dense trees' (from 1 false positive to 0)---all of which are habitat-related attributes. 
Similar trends can be observed on other classes as well such as in `Dense Trees' and `Low Plants' where we see higher true positive counts with WildSAT.

\paragraph{WildSAT is effective on zero-shot classification of habitat-related classes.} \cref{table:quant-zero-shot} shows zero-shot image classification results on classes from EuroSAT and So2Sat20k. WildSAT is geared towards classes related to nature and animal habitats, and thus see improvements in those classes (\eg `AnnualCrop', `SeaLake', `Water', `Low plants'). While WildSAT underperforms on anthropogenic classes (\eg `Highway', `Heavy industry') compared to models like TaxaBind and GRAFT, future work can integrate wildlife data and anthropogenic labels to create a more comprehensive and balanced representations.

\paragraph{Satellite image segmentation results.} \cref{table:segmentation-results} shows results of applying different ResNet50 encoders on satellite image segmentation. For each encoder, a convolution-based decoder is added while adopting the U-Net architecture~\cite{ronneberger2015u}. To accurately evaluate the features learned with and without WildSAT, we freeze the encoder and fine-tune only the decoder on the downstream dataset. \cref{table:segmentation-results} shows that WildSAT can provide rich representations leading to better segmentation performance. \cref{fig:segmentation} visualizes how WildSAT improves the segmentation outputs on Cashew1k and SAcrop3k. Further training on WildSAT leads to more accurate differentiation between the crop types.

\subsection{Additional Ablations}

\begin{table*}[!t]
    \setlength{\tabcolsep}{4pt}
    \small
    \begin{center}
    \begin{tabular}{l | c c c c | C{1.5cm} C{1.5cm} C{1.5cm} C{1.5cm} | c }
    \toprule
    \multicolumn{10}{c}{\textbf{ResNet50 SeCo}~\cite{manas2021seasonal_seco}} \\
    \midrule
    \multirow{2}{*}{Loss Terms} & \multirow{2}{*}{loc} & \multirow{2}{*}{env} & \multirow{2}{*}{text} & \multirow{2}{*}{img-a} & UCM & AID & RESISC45 & So2Sat20k & \multirow{2}{*}{Average} \\
     &  &  &  &  & \cite{yang2010bag_ucmdataset} & \cite{xia2017aid} & \cite{cheng2017remote_resisc45dataset} & \cite{zhu2020so2sat} &  \\
    \midrule
    Base Model &  &  &  &  & 86.1\% & 74.3\% & 80.2\% & 39.9\% & 70.1\% \\
    \midrule
    $\mathcal{L}_{\text{loc}}$ & \checkmark &  &  &  & 84.0\% & 76.2\% & 81.1\% & 43.5\% & 71.2\% \\
    $\mathcal{L}_{\text{loc}}$ & \checkmark & \checkmark &  &  & 84.1\% & 76.3\% & 83.0\% & 38.7\% & 70.5\% \\
    \midrule
    $\mathcal{L}_{\text{txt}}$ &  &  & \checkmark &  & 82.8\% & 74.4\% & 79.6\% & 39.5\% & 69.1\% \\
    $\mathcal{L}_{\text{txt}} + \mathcal{L}_{\text{loc}}$ & \checkmark &  & \checkmark &  & 84.3\% & 72.7\% & 78.5\% & 41.3\% & 69.2\% \\
    $\mathcal{L}_{\text{txt}} + \mathcal{L}_{\text{loc}}$ & \checkmark & \checkmark & \checkmark &  & 84.0\% & 75.8\% & 81.8\% & 40.0\% & 70.4\% \\
    \midrule
    $\mathcal{L}_{\text{img}}$ &  &  &  & \checkmark & 83.3\% & 75.0\% & 82.9\% & 46.0\% & 71.8\% \\
    $\mathcal{L}_{\text{img}}+\mathcal{L}_{\text{loc}}$ & \checkmark &  &  & \checkmark & 85.7\% & 77.9\% & 86.7\% & 46.2\% & 74.2\% \\
    $\mathcal{L}_{\text{img}}+\mathcal{L}_{\text{loc}}$ & \checkmark & \checkmark &  & \checkmark & 85.3\% & 77.1\% & 85.7\% & 48.2\% & 74.0\% \\
    \midrule
    $\mathcal{L}_{\text{img}}+\mathcal{L}_{\text{txt}}$ &  &  & \checkmark & \checkmark & 86.6\% & 77.0\% & 84.7\% & 48.6\% & 74.2\% \\
    $\mathcal{L}_{\text{img}}+\mathcal{L}_{\text{txt}}+\mathcal{L}_{\text{loc}}$ & \checkmark &  & \checkmark & \checkmark & 86.8\% & 78.1\% & 86.0\% & 49.0\% & 75.0\% \\
    $\mathcal{L}_{\text{img}}+\mathcal{L}_{\text{txt}}+\mathcal{L}_{\text{loc}}$ & \checkmark & \checkmark & \checkmark & \checkmark & 88.8\% & 79.6\% & 86.3\% & 46.0\% & \textbf{75.2\%} \\
    \bottomrule
    \toprule
    \multicolumn{10}{c}{\textbf{ViT-B/16 ImageNet}~\cite{deng2009imagenet}} \\
    \midrule
    \multirow{2}{*}{Loss Terms} & \multirow{2}{*}{loc} & \multirow{2}{*}{env} & \multirow{2}{*}{text} & \multirow{2}{*}{img-a} & UCM & AID & RESISC45 & So2Sat20k & \multirow{2}{*}{Average} \\
     &  &  &  &  & \cite{yang2010bag_ucmdataset} & \cite{xia2017aid} & \cite{cheng2017remote_resisc45dataset} & \cite{zhu2020so2sat} &  \\
    \midrule
    Base Model &  &  &  &  & 93.2\% & 84.4\% & 88.2\% & 41.8\% & 76.9\% \\
    \midrule
    $\mathcal{L}_{\text{loc}}$ & \checkmark &  &  &  & 95.2\% & 85.8\% & 89.3\% & 43.4\% & 78.4\% \\
    $\mathcal{L}_{\text{loc}}$ & \checkmark & \checkmark &  &  & 94.7\% & 86.2\% & 88.8\% & 44.2\% & 78.5\% \\
    \midrule
    $\mathcal{L}_{\text{txt}}$ &  &  & \checkmark &  & 96.1\% & 85.1\% & 88.9\% & 42.2\% & 78.1\% \\
    $\mathcal{L}_{\text{txt}} + \mathcal{L}_{\text{loc}}$ & \checkmark &  & \checkmark &  & 95.6\% & 86.5\% & 89.6\% & 39.6\% & 77.8\% \\
    $\mathcal{L}_{\text{txt}} + \mathcal{L}_{\text{loc}}$ & \checkmark & \checkmark & \checkmark &  & 95.1\% & 86.2\% & 89.7\% & 45.0\% & 79.0\% \\
    \midrule
    $\mathcal{L}_{\text{img}}$ &  &  &  & \checkmark & 97.1\% & 87.1\% & 91.5\% & 53.7\% & 82.3\% \\
    $\mathcal{L}_{\text{img}}+\mathcal{L}_{\text{loc}}$ & \checkmark &  &  & \checkmark & 97.1\% & 88.1\% & 91.7\% & 54.0\% & 82.7\% \\
    $\mathcal{L}_{\text{img}}+\mathcal{L}_{\text{loc}}$ & \checkmark & \checkmark &  & \checkmark & 96.8\% & 88.6\% & 92.1\% & 52.7\% & 82.6\% \\
    \midrule
    $\mathcal{L}_{\text{img}}+\mathcal{L}_{\text{txt}}$ &  &  & \checkmark & \checkmark & 96.9\% & 87.7\% & 92.0\% & 54.3\% & 82.7\% \\
    $\mathcal{L}_{\text{img}}+\mathcal{L}_{\text{txt}}+\mathcal{L}_{\text{loc}}$ & \checkmark &  & \checkmark & \checkmark & 97.1\% & 87.9\% & 91.9\% & 53.4\% & 82.6\% \\
    $\mathcal{L}_{\text{img}}+\mathcal{L}_{\text{txt}}+\mathcal{L}_{\text{loc}}$ & \checkmark & \checkmark & \checkmark & \checkmark & 97.5\% & 88.9\% & 93.0\% & 55.2\% & \textbf{83.6\%} \\
    \bottomrule
    \end{tabular}
    \vspace{-6pt}
    \caption{\textbf{Ablation of the various components of the WildSAT framework.} We ablate on SeCo~\cite{manas2021seasonal_seco}, a self-supervised pre-training method that applies contrastive learning on seasonal augmentations of images, and on ImageNet~\cite{deng2009imagenet}, a supervised pre-training on ImageNet. Through the different modalities in WildSAT, we can improve model-specific gaps.}
    \label{table:full-ablation}
    \end{center}
    \vspace{-6mm}
\end{table*}

\paragraph{WildSAT improves models by covering model-specific gaps.} \cref{table:full-ablation} displays an ablation study conducted on two different types of models: a ResNet50 SeCo~\cite{manas2021seasonal_seco} model and a ViT-B/16 ImageNet~\cite{deng2009imagenet} model. SeCo is pre-trained with a contrastive objective on time augmented satellite images (\ie satellite images from the same location, but from different seasons). This objective is similar to the $\mathcal{L}_{\text{img}}$ term (Eqn.~1 in the main paper) in the loss function of our WildSAT framework. Thus, we see from \cref{table:full-ablation}, that simply adding the image augmentation term (`img-a') only slightly improves the average performance across the downstream datasets (from 70.1\% to 71.8\%). This small improvement could be attributed to the additional examples related to habitats that are possibly not as well-represented in the SeCo dataset. However, if we add other modalities such as text and location (in addition to the satellite image augmentation), we see a larger improvement with an average performance of 74.2\% and 75.2\%, respectively. 
In contrast, an ImageNet pre-trained model benefits from satellite image augmentations ($\mathcal{L}_{\text{img}}$ or `img-a') since it was trained on a different domain.
Simply adding the image augmentation term improved average performance from 76.9\% to 82.3\%.
Further adding other modalities such as text and location also pushes the performance higher to 83.6\%.
These results support our hypothesis that WildSAT can complement and further improve different architectures by leveraging the different modalities.

\begin{table*}[!t]
    \setlength{\tabcolsep}{4pt}
    \small
    \begin{center}
    \begin{tabular}{c c | c | c c | C{1.5cm} C{1.5cm} C{1.5cm} C{1.5cm} | c }
    \toprule
    \multicolumn{10}{c}{\textbf{ResNet50 SeCo}~\cite{manas2021seasonal_seco}} \\
    \midrule
     \multicolumn{2}{c|}{No Model} & SatCLIP & \multicolumn{2}{c|}{SINR} & UCM & AID & RESISC45 & So2Sat20k & \multirow{2}{*}{Average} \\
     loc & env & loc & loc & env & \cite{yang2010bag_ucmdataset} & \cite{xia2017aid} & \cite{cheng2017remote_resisc45dataset} & \cite{zhu2020so2sat} &  \\
    \midrule
    &  &  &  &  & 86.6\% & 77.0\% & 84.7\% & 48.6\% & 74.2\% \\
    \midrule
    \checkmark &  &  &  &  & 87.3\% & 78.1\% & 85.5\% & 48.3\% & 74.8\% \\
    & \checkmark &  &  &  & 86.3\% & 76.3\% & 84.9\% & 47.2\% & 73.7\% \\
    \checkmark & \checkmark &  &  &  & 87.0\% & 77.4\% & 85.1\% & 48.4\% & 74.5\% \\
    &  & \checkmark &  &  & 86.0\% & 78.2\% & 85.5\% & 50.0\% & 74.9\% \\
    &  &  & \checkmark &  & 86.8\% & 78.1\% & 86.0\% & 49.0\% & 75.0\% \\
    &  &  & \checkmark & \checkmark & 88.8\% & 79.6\% & 86.3\% & 46.0\% & \textbf{75.2\%} \\
    \bottomrule
    \end{tabular}
    \vspace{-6pt}
    \caption{\textbf{Ablation of the location encoder.} These runs assume both the text ($\mathcal{L}_{\text{txt}}$) and the image augmentation($\mathcal{L}_{\text{img}}$) are already part of the model, which implicitly uses location (since images and text are matched based on location). We ablate the explicit addition of location as an input through different location encoders. We explore using no model (\ie directly just using the latitude/longitude or environmental covariates), SatCLIP~\cite{klemmer2023satclip}, and SINR~\cite{cole2023spatial_sinr}. The last row of the table corresponds to our WildSAT setup.}
    \label{table:location-encoder-ablation}
    \end{center}
    \vspace{-6mm}
\end{table*}

\paragraph{Location encoder ablation.} \cref{table:location-encoder-ablation} shows an ablation study conducted on our choice of the location encoder. We use a ResNet50 SeCo encoder as the base model, and report accuracy on downstream classification datasets. All rows in the table use WildSAT with satellite images and text ($\mathcal{L}_{\text{img}} + \mathcal{L}_{\text{txt}}$), which are matched based on the location---\ie location is implicitly used in all the results, and we ablate which encoder to use for explicitly including location as an input. We replace the location encoder in our WildSAT framework with one of the following: no model (\ie use the position encoded latitude and longitude and/or the raw environmental covariates vector), SatCLIP~\cite{klemmer2023satclip}, or SINR~\cite{cole2023spatial_sinr}. We use the SatCLIP pre-trained location encoder that takes the latitude and longitude as an input. For SINR, we explore the two variants of using (1) just the location (`loc') or (2) both the location (`loc') and environmental covariates (`env'). We find that the best average performance uses SINR with both the location and environmental covariates.

\paragraph{Parameter efficient fine-tuning (PEFT) preserves out-of-domain pre-training representations.} \cref{table:peft-results} displays the effect of fine-tuning all parameters of a given base model compared to fine-tuning specific layers (\ie applying PEFT). We compare the effect on out-of-domain pre-trained models (\eg ImageNet~\cite{recht2019imagenet}, CLIP~\cite{radford2021learning_clip}), and in-domain pre-trained models (\eg SatlasNet~\cite{bastani2023satlaspretrain}, SeCo~\cite{manas2021seasonal_seco}). 
We find that out-of-domain pre-trained models have better downstream performance by applying scale and shift fine-tuning~\cite{frankle2021training,lian2022scaling}, or by applying DoRa~\cite{mao2024dora}.
By fine-tuning specific layers, the models retain some of the original representations learned from the pre-training (\ie ImageNet or CLIP) so that performance does not deteriorate compared to the base models.
This has a significant impact on large models such as ViT, since fine-tuning all weights alters many parameters and risks shifting them in suboptimal directions.
On the other hand, while applying PEFT for in-domain pre-trained models SatlasNet~\cite{bastani2023satlaspretrain} and SeCo~\cite{manas2021seasonal_seco} improves performance compared to the base model, we see better performance when directly fine-tuning all the layers.
This may be because the model has already undergone pre-training on satellite images, making additional pre-training on similar data from WildSAT result in a non-disruptive shift.

\begin{table}[!t]
    \setlength{\tabcolsep}{4pt}
    \small
    \begin{center}
    \begin{tabular}{l l c | c c c c c c c }
    \toprule
    & \multirow{2}{*}{Encoder} & \multirow{2}{*}{PEFT}  & UCM & AID & RESISC45 & FMoW & EuroSAT & So2Sat20k  & BEN20k \\
     & & &\cite{yang2010bag_ucmdataset} & \cite{xia2017aid} & \cite{cheng2017remote_resisc45dataset} & \cite{christie2018functional_fmow} & \cite{helber2019eurosat} & \cite{zhu2020so2sat}  & \cite{sumbul2019bigearthnet} \\
    \midrule
    ResNet50 & ImageNet1K~\cite{recht2019imagenet} &             & 91.3\% & 82.0\% & 85.6\% & 42.1\% & 95.0\% & 47.0\% & 56.1\%\\
    ResNet50 & ImageNet1K~\cite{recht2019imagenet} & \checkmark & 93.6\% & 86.7\% & 90.1\% & 46.0\% & 96.0\% & 46.6\% & 57.5\%\\
    \midrule
    ViT-B/16 & CLIP~\cite{radford2021learning_clip} &           & 82.1\% & 71.0\% & 75.3\% & 34.9\% & 93.4\% & 50.4\% & 49.0\%\\
    ViT-B/16 & CLIP~\cite{radford2021learning_clip} & \checkmark & 96.3\% & 88.0\% & 93.0\% & 53.6\% & 97.1\% & 49.7\% & 59.1\%\\
    \midrule
    \midrule
    ResNet50 & SatlasNet~\cite{bastani2023satlaspretrain} &             & 90.1\% & 79.4\% & 85.4\% & 42.4\% & 95.4\% & 44.8\% & 56.4\%\\
    ResNet50 & SatlasNet~\cite{bastani2023satlaspretrain} & \checkmark & 86.9\% & 76.8\% & 82.0\% & 35.7\% & 94.1\% & 41.6\% & 51.6\% \\
    \midrule
    ResNet50 & SeCo~\cite{manas2021seasonal_seco} &            & 88.8\% & 79.6\% & 86.3\% & 42.8\% & 95.5\% & 46.0\% & 57.3\% \\
    ResNet50 & SeCo~\cite{manas2021seasonal_seco} & \checkmark & 86.7\% & 77.3\% & 83.2\% & 37.3\% & 94.0\% & 44.4\% & 54.8\% \\
    \bottomrule
    \end{tabular}
    \vspace{-6pt}
    \caption{\textbf{Ablation of parameter efficient fine-tuning (PEFT) when applied with WildSAT.} Models pre-trained on out-of-domain datasets (\eg ImageNet, CLIP) that are fine-tuned with PEFT can perform better on downstream tasks by preserving original representations from the base model. In contrast, models pre-trained on in-domain datasets (\eg SatlasNet, SeCo) show limited improvement from PEFT since the fine-tuning is in the same domain as the pre-training (\ie satellite images)---fine-tuning all layers has better performance.}
    \label{table:peft-results}
    \end{center}
    \vspace{-8mm}
\end{table}

\subsection{Zero-shot Retrieval} 
\begin{figure*}[!b]
    \begin{center}
    \includegraphics[scale=0.34]{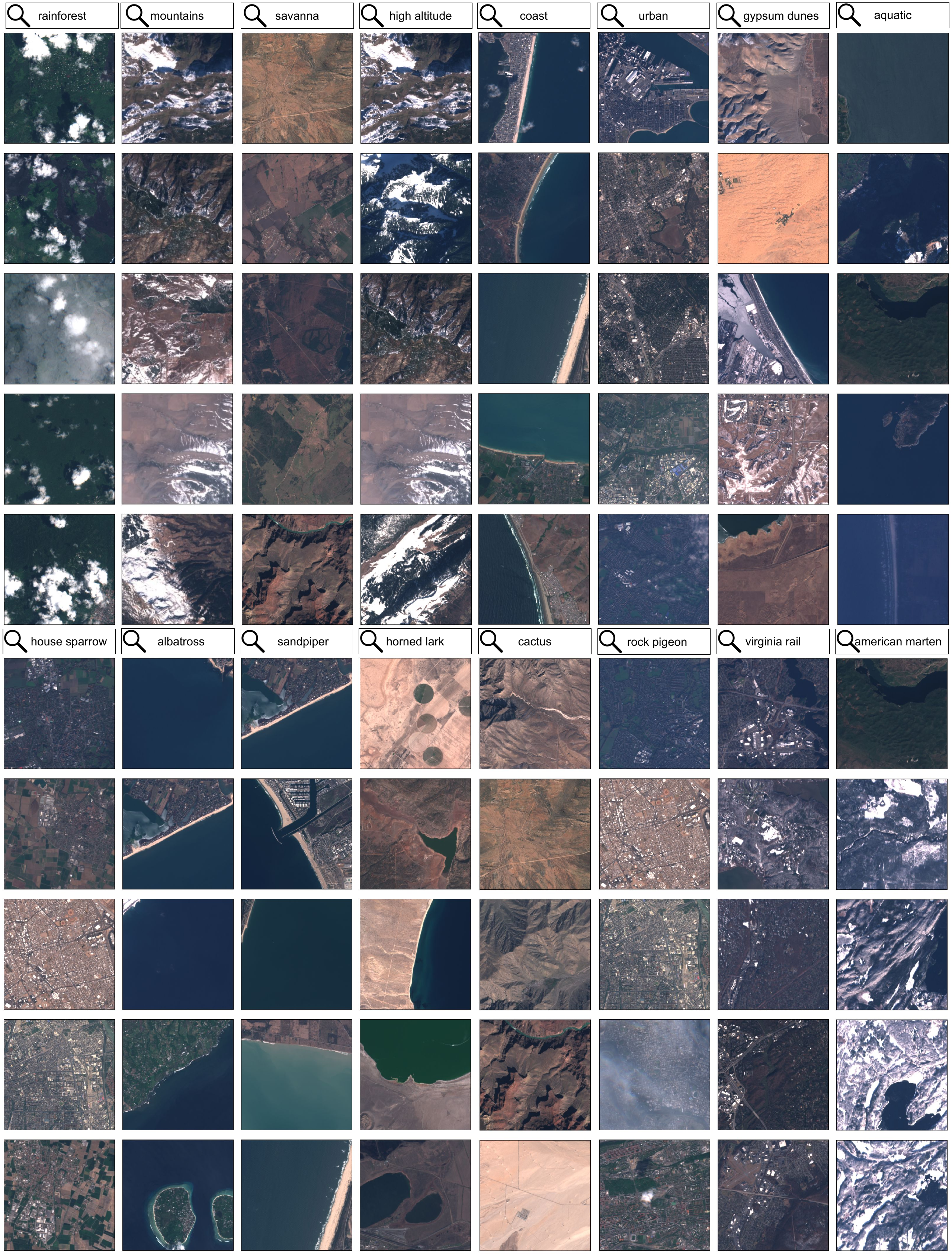}
    \end{center}
    \vspace{-16pt}
    \caption{\textbf{Additional zero-shot results for text-based satellite image retrieval}. }
    \label{fig:zero-shot-results-addtl}
    \vspace{-6mm}
\end{figure*}

In \cref{fig:zero-shot-results-addtl}, we display more zero-shot retrieval examples. The first row of examples demonstrates retrieval of general landscapes such as `rainforest' or `mountains'. The second row demonstrates retrieval of wildlife habitats. We enumerate each of the wildlife examples below including their expected habitats. All the enumerated habitats are consistent with the retrieved satellite images. \\
\noindent{}Description of the wildlife examples from the second set of rows in \cref{fig:zero-shot-results-addtl}:  
\begin{enumerate}
    \item `house sparrow' is a small, common bird typically found in urban areas.
    \item `albatross' is a large bird commonly found in the sea.
    \item `sandpiper' is a small bird that dwells in the coast.
    \item `horned lark' is a bird species found in open land such as on farmland, on prairies, and in deserts.
    \item `cactus' is a type of plant commonly found in the desert.
    \item `rock pigeon' is a bird commonly found in urban and residential areas.
    \item `virginia rail' is a bird found in freshwater and brackish marshes, and sometimes salt marshes in winter.
    \item `american marten' is a North American mammal that is found in forests, and broadly distributed in North America from Alaska and Canada to New York. 
\end{enumerate}

